%% file: acl_latex.tex
\pdfoutput=1
\documentclass[11pt]{article}

\usepackage[table]{xcolor}
\usepackage[final]{acl}

\usepackage{times}
\usepackage{latexsym}
\usepackage{caption}
\usepackage{booktabs}
\usepackage{multirow} 
\usepackage{makecell}
\usepackage{graphicx}
\usepackage{tcolorbox}
\usepackage{arydshln}
\usepackage{subcaption}
\usepackage{amsmath} 
\usepackage{kotex}
\usepackage{subcaption}
\usepackage{siunitx}
\usepackage{pifont}
\usepackage{cleveref}

\usepackage{adjustbox}
\usepackage{tabularx}

\usepackage[T1]{fontenc}
\usepackage[utf8]{inputenc}

\newcolumntype{L}[1]{>{\raggedright\arraybackslash}p{#1}}
\newcolumntype{C}[1]{>{\centering\arraybackslash}p{#1}}

\usepackage{microtype}
\usepackage{inconsolata}
\definecolor{gptPurple}{HTML}{D4C5F9}
\definecolor{pastelLavender}{HTML}{D4C5F9}
\definecolor{pastelPink}{HTML}{FFD1DC}
\definecolor{pastelBlue}{HTML}{C7E8FD}
\definecolor{pastelGreen}{HTML}{D1F2A5}
\definecolor{pastelMint}{HTML}{CFE7CC}
\definecolor{pastelOlive}{HTML}{DDE8D2}

\newcommand{\todo}[1]{}
\renewcommand{\todo}[1]{{\color{red} TODO{#1}}}

\title{CheckEval: A reliable LLM-as-a-Judge framework \\ for evaluating text generation using checklists}

\author{
 \textbf{Yukyung Lee\textsuperscript{1}}\quad
 \textbf{Joonghoon Kim\textsuperscript{2}}\quad
 \textbf{Jaehee Kim\textsuperscript{3,\#}}\quad
 \textbf{Hyowon Cho\textsuperscript{4,\#}}\quad
 \textbf{Jaewook Kang\textsuperscript{5}}\\
 \textbf{Pilsung Kang\textsuperscript{3,$\dagger$}}\quad
 \textbf{Najoung Kim\textsuperscript{1,$\dagger$}}
\\
\\
 \textsuperscript{1}Boston University\quad
 \textsuperscript{2}SK Telecom\quad
 \textsuperscript{3}Seoul National University\quad
 \textsuperscript{4}KAIST\quad
 \textsuperscript{5}NAVER
\\
{
\href{mailto:ylee5@bu.edu}{\texttt{ylee5@bu.edu}}\quad
\href{mailto:pilsung_kang@snu.ac.kr}{\texttt{pilsung\_kang@snu.ac.kr}}\quad
\href{mailto:najoung@bu.edu}{\texttt{najoung@bu.edu}}}
}

\begin{document}
\maketitle
\begingroup
\renewcommand\thefootnote{}\footnotetext{
    \textsuperscript{\#},\textsuperscript{†} Equal contribution. Our code is available at \url{https://github.com/yukyunglee/CheckEval}.} 
\endgroup

\input{sections/0_abstract}
\input{sections/1_introduction}
\input{sections/2_relatedwork}
\input{sections/3_method}
\input{sections/4_experiment_setup}
\input{sections/5_results}
\input{sections/6_human_validation}
\input{sections/7_analysis}
\input{sections/8_conclusion}
\input{sections/9_limitation}
\input{sections/10_acknowledgements}

\bibliography{custom}

\clearpage

\appendix
\input{sections/appendix}
\end{document}

%% file: sections/0_abstract.tex
\begin{abstract}
Existing LLM-as-a-Judge approaches for evaluating text generation suffer from rating inconsistencies, with low agreement and high rating variance across different evaluator models. We attribute this to subjective evaluation criteria combined with Likert scale scoring in existing protocols. To address this issue, we introduce CheckEval, a checklist-based evaluation framework that improves rating reliability via decomposed binary questions. Through experiments with 12 evaluator models across multiple datasets, we first demonstrate that CheckEval strongly correlates with human judgments. More importantly, CheckEval dramatically improves the average agreement across evaluator models by 0.45 and reduces the score variance. CheckEval scores furthermore have the benefit of being more interpretable because it decomposes evaluation criteria into traceable binary decisions, allowing analyses of specific attributes driving quality judgments.
\end{abstract} 

%% file: sections/1_introduction.tex
\section{Introduction}
Evaluating text generation quality remains a major challenge in Natural Language Generation (NLG), particularly as Large Language Models (LLMs) continue to advance in their generative capabilities \cite{brown2020language, chowdhery2023palm, achiam2023gpt}. This is especially evident in tasks such as summarization, dialogue, and creative writing \cite{liu2023calibrating, kim2023prometheus, liu2023x}, where qualitative dimensions of the output are crucial yet difficult to measure systematically. Consequently, there is growing interest in developing evaluation methods that can effectively capture these aspects. These methods will ideally involve well-defined protocols that ensure reliability across different raters and tasks. In obtaining actual scores from such protocols, human evaluation remains the gold standard, but it is costly, time-consuming, and difficult to scale \cite{novikova-etal-2017-need, belz-etal-2020-disentangling}. While lexical overlap-based metrics such as ROUGE and BLEU \cite{lin2004rouge, papineni2002bleu} have been widely adopted for ease of automation, they align poorly with human judgments, calling for alternatives that better approximate human evaluation. 

Recent work has explored the use of LLM-as-a-Judge as a scalable alternative, leveraging LLMs to assess generated text directly \cite{zheng2023judging}. This paradigm has evolved through various approaches: single-turn prompting \cite{liu2023gpteval, fu2023gptscore}, meta-evaluator training \cite{kim2023prometheus, wu-etal-2024-instructeval}, and even more sophisticated methods like multi-agent debate \cite{chan2024chateval, kim-etal-2024-debate}. 
However, these methods often rely on subjective evaluation protocols that require evaluators to assign holistic scores without clear decision criteria. For example, evaluators are typically asked to rate text on a Likert scale from 1 to 5 (higher is better) on dimensions such as coherence, consistency, fluency, and relevance. While Likert scales are useful for capturing ordinal relationships in human evaluation, they face two key challenges when applied to LLM-based evaluator models.
First, current LLMs are known to struggle with subjective criteria in Likert-scale evaluations, in particular showing difficulty in differentiating between high-quality texts \cite{li2019acute, stureborg2024large}.  
Second, evaluation results are highly sensitive to the choice of evaluator models. These lead to low \textit{inter-evaluator agreement} (IEA),\footnote{This is equivalent to Inter-Annotator Agreement (IAA) in human evaluation \cite{artstein2017inter}, but we use the term IEA in this paper to make it clear that the agreement we are aiming to improve is agreement between evaluator models, rather than between human raters providing the gold evaluation.} which we define as the agreement among evaluator models (of similar capacity), as well as high variance in evaluation results \cite{stureborg2024large}. Yet, previous LLM-as-a-Judge approaches have overlooked these issues \cite{gao2024analyzing}. 

\begin{figure*}[ht]
    \centering
    \includegraphics[width=\linewidth]{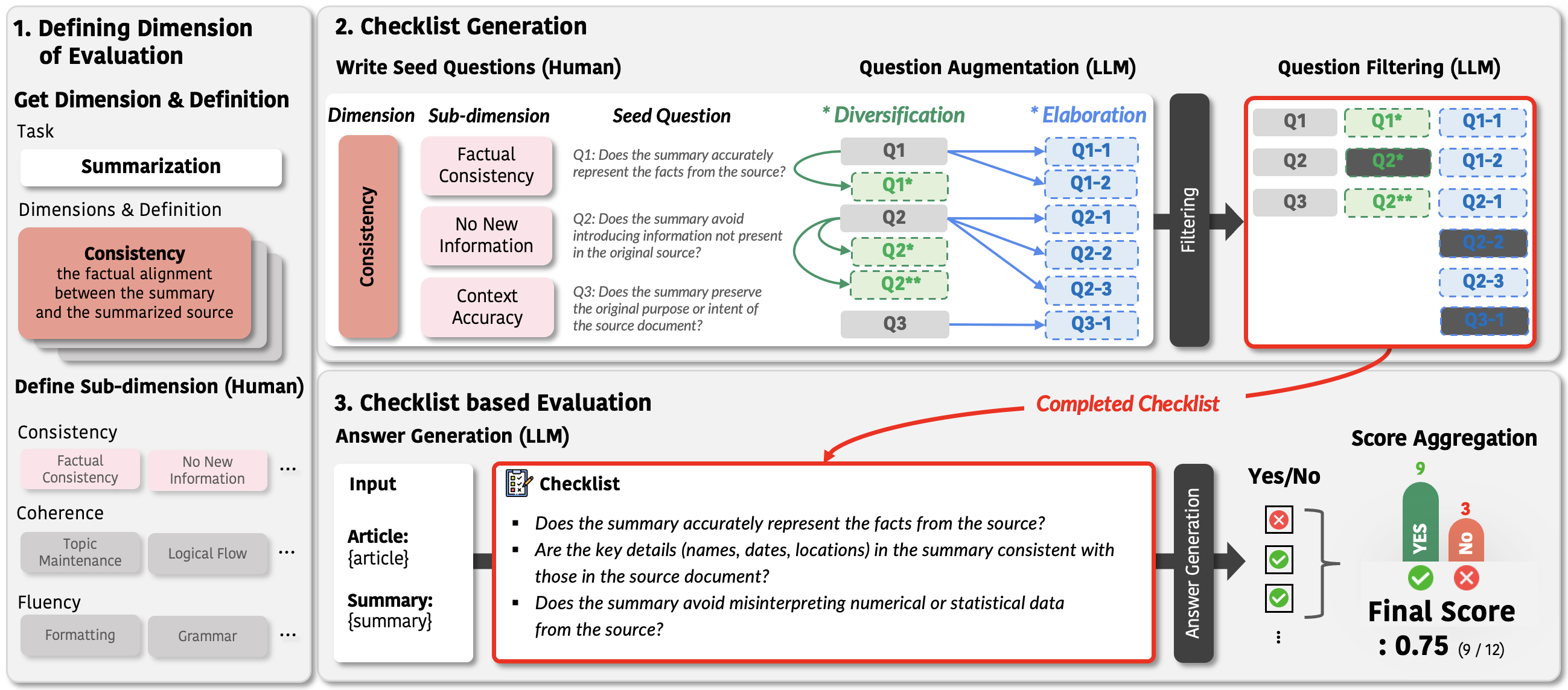}
    \caption{Overall process of CheckEval. CheckEval consists of three stages: (1) Defining Dimensions of Evaluation, where humans select specific dimensions and define sub-dimensions; (2) Checklist Generation, which incorporates two augmentation methods—question diversification (green) and elaboration (blue); and (3) Checklist-based Evaluation, where the model responds to the checklist with yes/no answers.}
    \label{fig:overview}
\end{figure*}

To address these challenges, we introduce \textbf{CheckEval}, a reliable evaluation framework that decomposes evaluation criteria to target fine-grained qualitative dimensions and turns them into a checklist.\footnote{Our checklist concept is inspired by \citet{ribeiro-etal-2020-beyond}, who proposed checklist-based testing for NLP models.} Inspired by recent advances in fine-grained decomposition of evaluation \cite{liu-etal-2023-revisiting, min-etal-2023-factscore}, our framework breaks down evaluation into discrete Boolean questions. This decomposition simplifies each individual evaluation question and clarifies the rationale behind evaluation decisions.
CheckEval addresses key limitations of existing methods in two ways. First, it improves explainability by tracking how specific criteria are met, making evaluation decisions more explicit and reducing ambiguity. Second, it enhances consistency through structured binary responses, which improve IEA and reduce variability. Importantly, CheckEval maintains competitive correlation with human evaluation while achieving these improvements. These improvements are verified through comprehensive experiments across 12 different LLM-based evaluator models of varying sizes, including both open and closed-source models, on multiple datasets. 
The main contributions of this study can be summarized as follows:
\begin{itemize}
\item We introduce CheckEval, a fine-grained evaluation framework leveraging a Boolean QA checklist to address the rating consistency issues with existing LLM-as-a-Judge methods for NLG evaluation.
\item Experiments across 12 LLMs and multiple datasets demonstrate significant improvements in correlation with human evaluation compared to Likert-based approaches like G-Eval \cite{liu2023gpteval} and SEEval \cite{wu-etal-2025-seeval}.
\item CheckEval shows reduced sensitivity to the choice of evaluator models, leading to more consistent evaluation results with lower variance and higher IEA.
\end{itemize} 

%% file: sections/2_relatedwork.tex
\section{Related Work}

\subsection{LLM-as-a-Judge} 
Traditional NLG evaluation metrics like ROUGE and BLEU show clear limitations due to their reliance on reference texts \cite{perceptionscore}. With advances in LLMs, researchers have explored LLM-as-a-Judge, where an LLM evaluates texts based on specified criteria, formalized as $F(\text{subject}, \text{criteria}) \to \text{result}$ \cite{li2024evaluating}. LLM-as-a-Judge can be categorized into pairwise and pointwise evaluation approaches \cite{gu2024survey}. Pairwise evaluation \cite{zheng2023judging, qin-etal-2024-large} compares two outputs to determine relative preference but is computationally expensive as comparisons scale exponentially. In contrast, pointwise evaluation \cite{liu2023gpteval, fu2023gptscore} assigns scores to individual outputs, allowing for absolute scaling. However, existing pointwise evaluation protocols often lack granularity, assigning a single numeric score to each dimension of evaluation. If the specified dimensions of evaluation are too broad (e.g., \texttt{fluency}), this may lead to inconsistencies in judgments because many factors could influence the quality along the target dimension. CheckEval falls in to the category of pointwise evaluation but addresses its limitations by adopting a finer-grained Boolean QA Checklist.\footnote{Recent studies \cite{wu2024meta, wang2024self} use LLM-as-a-Judge as a reward signal in alignment training with RLHF \cite{ouyang2022training}. However, this approach primarily aims to optimize model training rather than enhance evaluation robustness and explainability. Our work focuses on improving evaluation frameworks, and integrating evaluation signals into model training is beyond our scope.}

\subsection{Decompositional Approaches}

Decomposing complex information into minimal units to simplify tasks have been explored in various areas of NLP \cite{kamoi2023wice, chen2022generating, wright2022generating, krishna2023longeval, nenkova2004evaluating, liu-etal-2024-hd}. Recent studies have shown that breaking down content into atomic units reduces subjectivity in factual consistency judgment \cite{liu-etal-2023-revisiting, min-etal-2023-factscore}. Atomic units represent elementary information that cannot be further divided. Similarly, CheckEval decomposes evaluation criteria into fine-grained Boolean QA Checklists to enhance clarity and reduce ambiguity in the evaluation process.

\subsection{Reliability of Evaluation}
Reliability is an important yet often overlooked component of evaluation. Many LLM-as-a-Judge methods focus only on correlation with human scores, often neglecting consistency and stability across different LLMs. Recent studies have highlighted several reliability concerns. \citet{xiao-etal-2023-evaluating-evaluation} demonstrate that LLMs fail to reliably assess subtle quality differences in text. Similarly, \citet{DBLP:journals/corr/abs-2406-18403} find these models often assign highly variable ratings to identical inputs. Furthermore, IEA remains low across models, compromising evaluation reliability \cite{stureborg2024large}. Our work addresses these issues by evaluating not only correlation but also IEA and score variance across evaluator models, showing that CheckEval improves reliability across diverse LLMs.

%% file: sections/3_method.tex
\section{Method}
CheckEval consists of three stages, (1) Defining Dimensions of Evaluation, (2) Checklist Generation, and (3) Checklist-Based Evaluation, as shown in Figure \ref{fig:overview}.
The framework translates high-level evaluation criteria into a Boolean QA checklist, each question in the checklist expecting a binary (yes/no) response. This format improves clarity and alleviates ambiguity compared to Likert-scale scoring (discussed further in Section~\ref{subsec:case-study}).

\subsection{Defining Dimensions of Evaluation}
The first stage defines the dimensions of text quality (e.g., \texttt{consistency}, \texttt{fluency}) to be evaluated by either adopting predefined dimensions from benchmarks or specifying custom dimensions for the task. 
For each dimension, we then define sub-dimensions that break down the high-level dimensions further into distinct and detailed components. The sub-dimensions are grounded in the original definitions of the dimensions from benchmark datasets and can also also informed by related work \cite{liu-etal-2023-revisiting, laban-etal-2023-summedits, tang-etal-2019-topic}. For instance, the original SummEval paper proposes that \texttt{fluency} in summarization should include sub-dimensions such as \texttt{formatting, grammar, completeness}, and \texttt{readability}. 

Sub-dimensions must be carefully designed to align with benchmark definitions and to prevent inconsistencies with the intended evaluation criteria. While LLMs can be used to automate the generation of sub-dimensions and questions, we found that fully relying on them often led to misalignment with the criteria defined by the benchmark (e.g., conflating \texttt{coherence} and \texttt{fluency}). This leads to evaluation that is not grounded in the benchmark design, potentially producing incorrect assessments. To address this, we only allowed human-selected sub-dimensions in our work, following prior work that recommends human oversight as an effective way to maintain alignment with benchmark objectives \cite{szymanski2024comparing, pan-etal-2024-human}. 

\subsection{Checklist Generation}
\paragraph{Seed Question Writing} We create Boolean questions that correspond to the sub-dimensions defined in the first step. Each question requires a `yes’ or `no’ answer, where `yes’ indicates adherence to the evaluation criterion corresponding to the target sub-dimension. This binary format simplifies the judgment process, ensuring that evaluation criteria are explicitly defined and consistently applied \cite{laban-etal-2023-summedits, liu-etal-2023-revisiting}. This format also helps LLMs generate more reliable responses by constraining the answer space, minimizing response variability and reducing ambiguity. For example, the question \textit{``Are all words in the sentence spelled correctly?''} elicits a clearer and more direct response than a more open-ended alternative like \textit{``How well does the sentence adhere to or deviate from standard grammar rules?''}.
\paragraph{Question Augmentation} Manually designing a comprehensive set of evaluation questions would be ideal for ensuring a high-quality checklist. However, this approach faces scalability limitations, making it impractical to generate a sufficiently large and diverse set of questions for evaluation. This challenge becomes even more significant when extending to individual application scenarios, each requiring its own comprehensive set of questions. To this end, we expand the seed questions using LLMs, enhancing both the diversity and granularity of evaluation. Augmentation enables broader coverage while refining questions to capture a wider range of lexical and semantic variations. This process follows two strategies, each extending the coverage of seed questions. (1) \textit{Question Diversification} expands evaluation diversity by introducing variations that explore different perspectives of sub-dimensions and contexts of the seed question. 
(2) \textit{Question Elaboration} increases granularity by expanding the seed questions into more specific and detailed questions. To ensure that the augmented questions remain grounded in the seed questions, Question Diversification and Elaboration are performed independently rather than sequentially. For example, the seed question \textit{``Are all words in the sentence spelled correctly?''} can be expanded into \textit{``Are all sentences complete, with no fragments or missing components?''} (diversification) or specified into \textit{``Are proper nouns (names of people, places, etc.) spelled correctly?''} (elaboration). 
\paragraph{Question Filtering} LLM-based augmentation expands the question set, but it can also generate questions that do not fully align with the intended evaluation criteria. Some questions may reflect misinterpretations of dimension definitions or add unnecessary redundancy, which can affect evaluation reliability. To filter out such questions, we apply an LLM-based minimal filtering process that evaluates a combined pool of seed and augmented questions for each dimension. This filtering step applies three main criteria for retaining relevant questions: (1) \textit{alignment}, verifying that a `yes’ response to the question indicates higher quality; (2) \textit{dimension consistency}, confirming that the question adheres to the original definition of the evaluation dimension; and (3) \textit{redundancy removal}, eliminating semantically overlapping questions to avoid unnecessary repetition. While there is no direct metric to measure filtering effectiveness, we observe improved correlation with human judgments after filtering, suggesting that the filtering is functioning as intended. We further validated the quality of the checklist via a human study, where annotators scored the augmented and filtered questions (\Cref{sec:humaneval-checklist-validation}).

\subsection{Checklist-based Evaluation}
\label{subsec:checklist-based-evaluation}

In the final stage, LLMs evaluate the target text using the completed checklist (see Table \ref{tab:checklist-statistics-summeval} and \ref{tab:checklist-statistics-topicalchat} for the number of checklist questions and Table~\ref{tab:summary-dimensions} and \ref{tab:dialogue-dimensions} for the dimensions, sub-dimensions, and corresponding seed question for each dataset). To improve efficiency, we ask multiple questions simultaneously rather than asking each question separately. We compared single-question and multi-question inference in our pilot experiments and found no noticeable difference in performance. Therefore, we evaluated multiple questions together to reduce the computational cost. The questions are grouped by sub-dimensions, ensuring that related questions are presented together to aid model comprehension. For each question in the checklist, the LLM generates a `yes’ or `no’ response. The final quality score is computed as the proportion of `yes’ responses among all questions (e.g., 15 `yes’ out of 20 questions yields 0.75). We note that the final score is computed by uniformly weighting the checklist questions: each `yes' response contributes equally to the final score. We discuss an alternative weighting strategy in Appendix~\ref{sec:weighted question aggregation}. More implementation details about the evaluation process are described in Section~\ref{subsec:implementation-details}.

This approach enhances explainability by explicitly tracking how specific criteria are met, making evaluation decisions more interpretable without requiring additional rationale generation. Unlike existing LLM-as-a-Judge approaches such as G-Eval and SEEval (our main comparison points) that generate numerical scores without explanation (e.g., \textit{``Based on the conversation history, the corresponding context, and the response, here is the evaluation: `Naturalness': 2''}), the reasoning behind the evaluation score is easily traceable from the checklist responses.

%% file: sections/4_experiment_setup.tex
\section{Experimental Setup}
\subsection{Datasets and Metrics}
Following~\citet{liu2023gpteval}, We use three meta-evaluation benchmarks spanning various tasks to measure the effectiveness of CheckEval. \textbf{SummEval} \cite{fabbri2021summeval} is a benchmark designed for the meta-evaluation of summarization. SummEval includes human evaluations for each generated summary across four dimensions: \texttt{coherence}, \texttt{consistency}, \texttt{fluency}, and \texttt{relevance}. \textbf{Topical-Chat} \cite{gopalakrishnan2019topical} is a benchmark for meta-evaluating evaluation methods for knowledge-grounded dialogue systems.
Following \citet{zhong2022towards}, we evaluate our method using human ratings across four dimensions: \texttt{naturalness}, \texttt{coherence}, \texttt{engagingness}, and \texttt{groundedness}.
\textbf{QAGS} \cite{wang2020asking} is another widely used benchmark, but since it focuses solely on factual consistency in summarization, we only report the results in Appendix~\ref{sec:appendix-qags-experiments}. We report Pearson's $r$, Spearman's $\rho$, Kendall's $\tau$ on each benchmark. For SummEval, correlations are calculated at the sample-level (per summary), while for Topical-Chat, they are calculated at the turn-level (per conversational response).

\subsection{Baselines} 

We selected \textbf{G-Eval} \cite{liu2023gpteval} and \textbf{SEEval} \cite{wu-etal-2025-seeval} as our main baselines. G-Eval adopts chain-of-thought prompting \cite{wei2022chain} and a form-filling paradigm to generate evaluation scores on a Likert scale. We selected it based on three factors:
(1) its widespread adoption as a representative baseline in LLM-as-a-judge research, (2) the availability of publicly released prompts that facilitate reproducibility, and (3) its relatively simple setup that avoids confounding performance-enhancing techniques—such as prompt optimization (e.g., self-correction), training meta-evaluators, preference learning, or multi-agent frameworks. SEEval  follows a similar Likert-style scoring procedure to G-Eval but augments it with a self-explanation step, prompting the model to generate brief justifications before producing its rating. This strategy is intended to improve evaluation quality without additional training.

Like G-Eval and SEEval, CheckEval is also designed to rely solely on a binary checklist mechanism, without introducing additional optimization techniques beyond standard prompting. Although they are not apples-to-apples comparisons, we also include comparisons to several strong methods surveyed in \citet{gu2024survey} and \citet{Gao2024LLMbasedNE}, showing that CheckEval remains competitive even in light of more recent developments. Further details on the baseline implementations are provided in Appendix~\ref{subsec:baselines}.

\subsection{Models}
We test both open-source models of varying sizes and closed-source GPT models as evaluators. The models included in each category are as follows:\footnote{The links for each model are provided in Appendix \ref{sec:appendix-link}.}
(1) \textbf{Large models} (70--123B): \texttt{LLama3.1-70B}, \texttt{Mistral-Large} (123B), \texttt{Qwen2.5-72B}. (2) \textbf{Medium models} (22--32B): \texttt{Mistral-Small} (22B), \texttt{Gemma2-27B}, \texttt{Qwen2.5-32B}. (3) \textbf{Small models} (7--9B): \texttt{LLama3.1-8B}, \texttt{Gemma2-9B}, \texttt{Qwen2.5-7B}, (4) \textbf{GPT models}: \texttt{GPT-4-Turbo}, \texttt{GPT-4o}, \texttt{GPT-4o-mini} \cite{achiam2023gpt, dubey2024llama, jiang2023mistral, yang2024qwen2, team2024gemma}.

\subsection{Implementation Details}
\label{subsec:implementation-details}

Following prior work \cite{liu2023gpteval}, we set \texttt{temperature} = 0, \texttt{n} = 1, and fix the random seed for both G-Eval, SEEval and CheckEval. Additionally, We set \texttt{max\_length} to 20 for G-Eval as it generates a single score, 500 for SEEval following the original implementation and 200 for CheckEval as it needs to generate responses to multiple checklist questions. We used the original prompts provided by the authors of G-Eval and SEEval without any modifications. Example prompts for CheckEval are provided in the Appendix~\ref{sec:appendix-prompt}. We evaluated multiple questions in the checklist within a single prompt to enhance efficiency and practicality rather than evaluating each question individually, as discussed in Section~\ref{subsec:checklist-based-evaluation}. 

We used GPT-4o for both the question augmentation and filtering steps in the checklist generation stage. The total number of generated questions at each step is provided in Appendix \ref{sec:appendix-number of questions}. Our experiments on open-weight models were conducted using vLLM 0.6.3 \cite{kwon2023efficient} with four A100 GPUs (or eight A6000 GPUs). The API cost to evaluate the 1,600 SummEval samples was approximately \$66 with GPT-4 Turbo, \$22 with GPT-4o, and \$1.30 with GPT-4o-mini.

\begin{table}[t]
    \centering
    \setlength{\tabcolsep}{5pt}
    \renewcommand{\arraystretch}{1.2}
    \resizebox{0.9\columnwidth}{!}{%
        \footnotesize
        \begin{tabular}{p{2cm} p{1.8cm} >{\centering\arraybackslash}p{1cm} >{\centering\arraybackslash}p{1cm} >{\centering\arraybackslash}p{1cm} >{\centering\arraybackslash}p{1cm}}
            \toprule
            \multirow{2}{*}{\makecell[l]{\textbf{Model}\\ \textbf{} }}  
            & \multirow{2}{*}{\makecell[l]{\textbf{Evaluation}\\ \textbf{Methods}}} 
            & \multicolumn{2}{c}{\textbf{SummEval (Avg.)}} 
            & \multicolumn{2}{c}{\textbf{Topical-Chat (Avg.)}} \\
            & & $\rho$ & $\tau$ & $\rho$ & $r$ \\
            \midrule
            \multicolumn{6}{l}{\textbf{\textit{non-LLM-as-a-Judge}}} \\
            \midrule
            & ROUGE-L & 0.17 & 0.13 & 0.24 & 0.24 \\
            & BERTScore & 0.23 & 0.18 & 0.25 & 0.24 \\
            & MOVERScore & 0.47 & 0.38 & 0.22 & 0.24 \\
            & BARTScore & 0.19 & 0.15 & 0.29 & 0.29 \\
            & UniEval & 0.39 & 0.31 & 0.28 & 0.26 \\
            \midrule
            \multicolumn{6}{l}{\textbf{\textit{LLM-as-a-Judge}}} \\
            \midrule
            \cellcolor{pastelPink}\textbf{Llama3.1-70B}
            & G-Eval & 0.40 & 0.36 & 0.45 & 0.39 \\
            \cellcolor{pastelPink}
            & SEEval & 0.41 & 0.35 & 0.55 & 0.54 \\
            \cellcolor{pastelPink}
            & CheckEval & \textbf{0.46} & \textbf{0.40} & \textbf{0.57} & \textbf{0.57} \\
            \cdashline{2-6}
            \cellcolor{pastelPink}\textbf{Mistral-Large}
            & G-Eval & 0.52 & 0.47 & 0.64 & 0.62 \\
            \cellcolor{pastelPink}
            & SEEval & 0.54 & 0.50 & 0.64 & 0.63 \\
            \cellcolor{pastelPink}
            & CheckEval & \textbf{\underline{0.55}} & \textbf{\underline{0.48}} & \textbf{\underline{0.65}} & \textbf{\underline{0.65}} \\
            \cdashline{2-6}
            \cellcolor{pastelPink}\textbf{Qwen2.5-72B}
            & G-Eval & 0.43 & 0.39 & \textbf{0.62} & \textbf{0.61} \\
            \cellcolor{pastelPink}
            & SEEval & 0.47 & 0.41 & 0.60 & 0.60 \\
            \cellcolor{pastelPink}
            & CheckEval & \textbf{0.50} & \textbf{0.44} & 0.59 & 0.60 \\
            \midrule
            \cellcolor{pastelBlue}\textbf{Mistral-Small}
            & G-Eval & 0.18 & 0.16 & \textbf{0.58} & \textbf{0.52} \\
            \cellcolor{pastelBlue}
            & SEEval & 0.22 & 0.20 & 0.17 & 0.17 \\
            \cellcolor{pastelBlue}
            & CheckEval & \textbf{0.45} & \textbf{0.39} & 0.47 & 0.49 \\
            \cdashline{2-6}
            \cellcolor{pastelBlue}\textbf{Gemma2-27B}
            & G-Eval & 0.44 & 0.39 & 0.31 & 0.29 \\
            \cellcolor{pastelBlue}
            & SEEval & 0.44 & 0.39 & 0.41 & 0.44 \\
            \cellcolor{pastelBlue}
            & CheckEval & \textbf{0.51} & \textbf{0.44} & \textbf{0.53} & \textbf{0.52} \\
            \cdashline{2-6}
            \cellcolor{pastelBlue}\textbf{Qwen2.5-32B}
            & G-Eval & 0.50 & \textbf{0.45} & 0.46 & 0.38 \\
            \cellcolor{pastelBlue}
            & SEEval & 0.49 & 0.44 & 0.47 & 0.51  \\
            \cellcolor{pastelBlue}
            & CheckEval & \textbf{0.52} & 0.44 & \textbf{0.56} & \textbf{0.56} \\
            \midrule
            \cellcolor{pastelOlive}\textbf{Llama3.1-8B}
            & G-Eval & 0.24 & 0.21 & 0.11 & 0.09 \\
            \cellcolor{pastelOlive}
            & SEEval & 0.16 & 0.13 & 0.17 & 0.17 \\
            \cellcolor{pastelOlive}
            & CheckEval & \textbf{0.41} & \textbf{0.34} & \textbf{0.46} & \textbf{0.45} \\
            \cdashline{2-6}
            \cellcolor{pastelOlive}\textbf{Gemma2-9B}
            & G-Eval & 0.38 & 0.34 & 0.46 & 0.35 \\
            \cellcolor{pastelOlive}
            & SEEval & 0.49 & 0.40 & \textbf{0.49} & \textbf{0.50} \\
            \cellcolor{pastelOlive}
            & CheckEval & \textbf{0.43} & \textbf{0.37} & \textbf{0.49} & \textbf{0.50} \\
            \cdashline{2-6}
            \cellcolor{pastelOlive}\textbf{Qwen2.5-7B}
            & G-Eval & 0.41 & \textbf{0.38} & 0.45 & 0.39 \\
            \cellcolor{pastelOlive}
            & SEEval & 0.39 & 0.34 & \textbf{0.48} & 0.46 \\
            \cellcolor{pastelOlive}
            & CheckEval & \textbf{0.42} & 0.37 & \textbf{0.48} & \textbf{0.47} \\
            \midrule
            \cellcolor{pastelLavender}\textbf{GPT-4 Turbo} 
            & G-Eval & 0.51 & \textbf{0.46} & 0.59 & 0.58 \\
            \cellcolor{pastelLavender}
            & SEEval & 0.50 & \textbf{0.46} & 0.60 & 0.61 \\
            \cellcolor{pastelLavender}
            & CheckEval & \textbf{0.52} & \textbf{0.46} & \textbf{0.63} & \textbf{0.64} \\
            \cdashline{2-6}
            \cellcolor{pastelLavender}\textbf{GPT-4o}
            & G-Eval & 0.32 & 0.29 & 0.52 & 0.43 \\
            \cellcolor{pastelLavender}
            & SEEval & 0.39 & 0.37 & 0.56 & 0.47 \\
            \cellcolor{pastelLavender}
            & CheckEval & \textbf{0.50} & \textbf{0.44} & \textbf{0.64} & \textbf{0.63} \\
            \cdashline{2-6}
            \cellcolor{pastelLavender}\textbf{GPT-4o-mini}
            & G-Eval & 0.45 & 0.40 & 0.58 & 0.56 \\
            \cellcolor{pastelLavender}
            & SEEval & 0.46 & 0.41 & 0.57 & 0.56 \\
            \cellcolor{pastelLavender}
            & CheckEval & \textbf{0.49} & \textbf{0.42} & \textbf{0.59} & \textbf{0.59} \\
            \bottomrule
        \end{tabular}
    }
    \caption{Average correlation scores across dimensions on the benchmarks. For SummEval, we report sample-level $\rho$ and $\tau$. For Topical-Chat, we report turn-level $\rho$ and $r$. Colors indicate model groups: large (pink), medium (blue), small (green) and  GPT (purple). The best score per model category is bolded, and the highest overall score is marked with an underline.}
    \label{tab:merged-results}
\end{table}

%% file: sections/5_results.tex
\section{Results}

\subsection{Correlation with Human Evaluation}
Table \ref{tab:merged-results} shows the correlation between various evaluation methods and human judgments on the SummEval and Topical-Chat datasets (detailed correlation results for all dimensions are shown in Table~\ref{tab:main-corr-result-summeval}, \ref{tab:main-corr-result-topical-chat} and \ref{tab:qags-correlation} in the Appendix). We compare both non-LLM-as-a-Judge and LLM-as-a-Judge methods, with an emphasis on how CheckEval compares against G-Eval and SEEval across 12 LLMs.

Excluding MOVERScore, most non-LLM-as-a-Judge metrics exhibit very low correlation with humans. Among LLM-as-a-Judge methods, CheckEval consistently achieves higher correlation with human judgments than G-Eval and SEEval, with only a few exceptions of Qwen2.5 and Mistral-Small. These results suggest that CheckEval's fine-grained, checklist-based design more effectively captures subtle differences in text quality, leading to improved correlation with human judgments. When analyzing model sizes, large open-source models show strong performance, with Mistral-Large combined with CheckEval achieving the highest correlation among all models ($\rho=0.55$ on SummEval and $r=0.65$ on Topical-Chat). Even in medium- and small-sized models---where evaluation capacity tends to be weaker---CheckEval maintains its advantage over G-Eval. Notably, some medium-sized models perform particularly well on SummEval, achieving correlations comparable to larger models. For GPT models, CheckEval consistently yields stronger correlations than G-Eval and SEEval, particularly with GPT-4o. 

\begin{table}[t]
\centering
\setlength{\tabcolsep}{5pt}
\renewcommand{\arraystretch}{1.1}
\resizebox{0.95\columnwidth}{!}{%
\footnotesize
    \begin{tabular}{llcccc}
    \toprule
    \multirow{2}{*}{\makecell[l]{\textbf{Model}\\ \textbf{Group}}}  
    & \multirow{2}{*}{\makecell[l]{\textbf{Evaluation}\\ \textbf{Methods}}} 
    & \multicolumn{2}{c}{\textbf{SummEval (Avg.)}} 
    & \multicolumn{2}{c}{\textbf{Topical-Chat (Avg.)}} \\
    \cmidrule(lr){3-4} \cmidrule(lr){5-6}
    & & \textbf{$\alpha$} & \textbf{$\kappa$} 
    & \textbf{$\alpha$} & \textbf{$\kappa$} \\
    \midrule
    \multirow{2}{*}{All} & G-Eval & 0.09 & 0.19 & 0.06 & 0.34 \\
    & SEEval & 0.08 & 0.14 & 0.07 & 0.31 \\
    & CheckEval & \textbf{0.48} & \textbf{0.48} & \textbf{0.45} & \textbf{0.45} \\
    \midrule
    \multirow{2}{*}{Large} & G-Eval & 0.05 & 0.16 & 0.01 & 0.51 \\
    & SEEval & 0.06 & 0.19 & 0.55 & 0.61 \\
    & CheckEval & \textbf{0.67} & \textbf{0.67} & \textbf{0.67} & \textbf{0.67} \\
    \midrule
    \multirow{2}{*}{Medium} & G-Eval & 0.04 & 0.14 & 0.07 & 0.22 \\
    & SEEval & 0.09 & 0.13 & 0.06 & 0.34 \\
    & CheckEval & \textbf{0.56} & \textbf{0.56} & \textbf{0.50} & \textbf{0.50} \\
    \midrule
    \multirow{2}{*}{Small} & G-Eval & 0.06 & 0.10 & 0.04 & 0.16 \\
    & SEEval & 0.02 & 0.07 & 0.16 & 0.15 \\
    & CheckEval & \textbf{0.24} & \textbf{0.24} & \textbf{0.17} & \textbf{0.17} \\
    \midrule
    \multirow{2}{*}{GPT} & G-Eval & 0.08 & 0.20 & 0.04 & 0.50 \\
    & SEEval & 0.13 & 0.32 & 0.12 & 0.51 \\
    & CheckEval & \textbf{0.56} & \textbf{0.56} & \textbf{0.54} & \textbf{0.54} \\
    \midrule
    \multirow{2}{*}{Top-3$^\ast$} & G-Eval & 0.07 & 0.23 & 0.03 & 0.56 \\
    & SEEval & 0.09 & 0.19 & 0.06 & 0.34 \\
    & CheckEval & \textbf{0.65} & \textbf{0.65} & \textbf{0.57} & \textbf{0.57} \\
    \bottomrule
    \end{tabular}}
\caption{Inter-evaluator agreement (IEA) results for SummEval and Topical-Chat, comparing G-Eval, SEEval and CheckEval across different model groups. Top-3 refers to the three models with the highest correlation to human judgments ($ ^\ast$ see Appendix~ \ref{subsec:top3models} for the list of top-3 models for each evaluation method). The best score per model category is bolded.}
\label{tab:iea-combined}
\end{table}

\subsection{Inter-evaluator Agreement (IEA)}
\label{subsec:agreement-analysis}

Table \ref{tab:iea-combined} compares the IEA of G-Eval, SEEval and CheckEval on the SummEval and Topical-Chat datasets. We measure IEA using Krippendorff's $\alpha$ and Fleiss' $\kappa$, treating different LLMs within the same group (large, medium, small, GPT) as annotators. While correlation with human judgments is a main metric in LLM-as-a-Judge, \textbf{high correlation alone does not guarantee reliability}. Reliability is a desirable property for evaluation methods, as it ensures that different evaluator models (of similar capacity) assign similar scores/rating to the same input. This reliability is critical yet overlooked in existing frameworks.

Both G-Eval and SEEval demonstrate this limitation. They achieve fairly good correlation with human judgments but show much lower IEA in general. Table~\ref{tab:iea-combined} shows a clear gap between the IEA of G-Eval and SEEval and IEA of CheckEval, particularly for the Large and Top-3 models. 
This indicates inconsistent scoring across different LLM evaluator models of similar capacity. We speculate that existing protocols like G-Eval's mainly lend themselves to inconsistencies in the following two ways: (1) the evaluation dimensions adopted encompass multiple distinct fine-grained criteria, making it difficult for LLMs to generate a consistent holistic score, and (2) adjacent Likert scale scores lack clear distinctions (e.g., 3 vs. 4) and are not calibrated well across models \cite{laban-etal-2023-summedits}.

\begin{figure}[t]
    \centering
\includegraphics[width=\columnwidth]{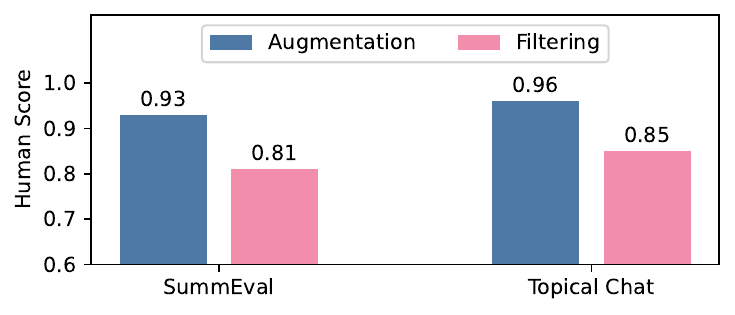}
    \caption{Human validation scores for the checklist generation process, averaged across all dimensions on both SummEval and Topical-Chat. `Augmentation' refers to the percentage of augmented questions that fulfilled the specified quality criteria, and `Filtering' refers to the percentage for filtered questions.}
    \label{fig:checklist-validation}
\end{figure}

CheckEval's fine-grained checklist approach improves upon this limitation greatly. For the large models, CheckEval achieves best IEA scores of 0.67 ($\alpha$ and $\kappa$), on SummEval, which is comparable to IEA among human raters ($\kappa \approx 0.7$) \cite{fabbri2021summeval}, and 0.67 ($\alpha$ and $\kappa$) on Topical-Chat. Crucially, CheckEval maintains both high correlation and  IEA across different LLMs and tasks.  These results demonstrate that CheckEval provides a more reliable evaluation than G-Eval and SEEval (See Table \ref{tab:iea-full result-SummEval} and \ref{tab:iea-full result-topicalchat} for a detailed per-dimension IEA). We furthermore show that this improvement in IEA is not solely due to the format of the output (Likert vs. binary) in Appendix~\ref{binarizing-likert}.

%% file: sections/6_human_validation.tex
\section{Human Validation}
We conducted two distinct human evaluation studies to validate our approach: (1) an assessment of our automated checklist generation process, and (2) a direct comparison between LLM and human scores using the CheckEval protocol.

\subsection{Validation of Checklist Generation Process}
\label{sec:humaneval-checklist-validation}

\begin{table}[t]
\centering
\setlength{\tabcolsep}{7pt}
\renewcommand{\arraystretch}{1.3}
\resizebox{0.9\columnwidth}{!}{%
\footnotesize
\begin{tabular}{llcc} 
\toprule
\multirow[t]{2}{*}{\textbf{Model}} & \multirow{2}{*}{\begin{tabular}[c]{@{}l@{}}\textbf{Evaluation}\\\textbf{Method}\end{tabular}} & \multicolumn{2}{c}{\textbf{SummEval (Avg.)}} \\ 
\cmidrule(l){3-4} & & $\rho$& $\tau$ \\ 
\midrule
\multirow[t]{6}{*}{\textbf{Mistral-Large}} 
& CheckEval                        & 0.5486          & 0.4797         \\
& \textbf{CheckEval $^\#$}    & 0.5486          & 0.4797         \\ 
\cmidrule(l){2-4}
& &\multicolumn{2}{c}{\textbf{Topical-Chat (Avg.)}}                                         \\ 
\cmidrule(l){3-4} & & $\rho$& $r$ \\ 
\cmidrule(l){2-4}
& CheckEval                   & 0.6451          & 0.6453         \\
& \textbf{CheckEval $^\#$}    & 0.6443          & 0.6412         \\ 
\cmidrule(l){2-4}
& & \multicolumn{2}{c}{\textbf{QAGS (Avg.)}}\\ 
\cmidrule(l){3-4} & & $r$& $\rho$ \\ 
\cmidrule(l){2-4}
& CheckEval                    & 0.6681          & 0.6558         \\
& \textbf{CheckEval $^\#$}     & 0.6680          & 0.6558         \\
\bottomrule
\end{tabular}}
\caption{Effect of applying additional human filtering to Mistral-Large. \# indicates that filtering was applied.}
\label{tab:checklist-validation}
\end{table}

To verify that each stage of the checklist generation process worked as intended, we conducted an additional human evaluation focused on checklist quality. This evaluation validates the augmentation stage (seed questions, augmented questions), and filtering stage (seed questions, filtered questions) on both the SummEval and Topical-Chat datasets. Human evaluators are tasked with assessing each question on a binary (yes/no) basis, determining whether it satisfies the augmentation and filtering criteria. Figure~\ref{fig:checklist-validation} shows the average scores derived from the checklist validation evaluation for both the SummEval and Topical-Chat datasets. The augmentation stage consistently achieves very high average scores across both datasets (above 90\%), which suggests that the question augmentation process of CheckEval is highly effective. The filtering stage yields slightly lower scores but remains competitive. We observed that annotators often expected 1–2 additional questions per dimensions to be filtered. Comments from annotators suggest that these questions were mostly semantically overlapping questions that the filter failed to capture.

To test whether removing these remaining questions would affect evaluation results, we conducted a follow-up experiment by applying an additional human-curated filtering step. We used Mistral-Large, the best-performing model, for this experiment. As shown in As shown in Table~\ref{tab:checklist-validation}, the correlation scores after applying the additional filtering were extremely similar to the original results, with only minor drops. This indicates that removing one or two additional questions per evaluation dimension does not meaningfully impact the evaluation behavior, suggesting that CheckEval’s automatic filtering is functioning effectively in practice.

\subsection{Validation of CheckEval Protocol}
\label{subsec:human-validation}

\begin{table}[t]
\centering
\renewcommand{\arraystretch}{1.2}
\resizebox{0.95\columnwidth}{!}{%
\footnotesize
\begin{tabular}{lcc}
\toprule
\textbf{\textit{Correlation}} & $\rho$ & $\tau$\\ 
\midrule
\phantom{00}\texttt{\textbf{Mistral-large}} (C) vs. Humans (C)  & 0.73$^{\ast\ast}$ & 0.58$^{\ast\ast}$\\
\phantom{00}\texttt{\textbf{Qwen2.5-72B}} (C) vs. Humans (C)& 0.72$^{\ast\ast}$ & \phantom{0}0.59$^{\ast\ast\ast}$\\
\phantom{00}\texttt{\textbf{Llama3.1-70B}} (C) vs. Humans (C)& 0.73$^{\ast\ast}$ & 0.58$^{\ast\ast}$\\
\cdashline{1-3}
\phantom{00}Humans (L) vs. Humans (\text{C}) & 0.69$^{\ast\ast}$ & \phantom{0}0.54$^{\ast\ast\ast}$ \\
\midrule
\textbf{\textit{Agreement}} \textit{(dim: Relevance)} & \# Annotators & $\kappa$   \\ 
\midrule
\phantom{00}Humans& 3 & 0.53  \\
\phantom{00}LLMs (Large) \& Humans & 6 & 0.49  \\
\bottomrule
\end{tabular}}
\caption{Human validation of the CheckEval protocol on SummEval. \textbf{C} denotes CheckEval, \textbf{L} denotes Likert (original SummEval Score). We use the LLM results from the large model group. ($^{\ast\ast}$: $p<.01$, $^{\ast\ast\ast}$: $p<.001$)}
\label{tab:humaneval}
\end{table}

To further assess the validity of CheckEval protocol, we asked human annotators to manually apply the same checklist. We then used these human-generated scores to perform two analyses: a correlation analysis against scores from LLMs, and an inter-rater agreement analysis (Table~\ref{tab:humaneval}). Details of the human validation setup are provided in Appendix~\ref{sec:humaneval-setting}.

\paragraph{Correlation} We sampled 20 summaries from the SummEval dataset. The random sampling was stratified based on original annotation scores to ensure balanced coverage of a wide range of quality levels.
Three annotators evaluated each summary using the checklist, which contains approximately 25 binary (yes/no) questions per evaluation dimension. This resulted in roughly 2,000 binary annotations per annotator. For each summary, we aggregated checklist scores by summing the number of ‘yes’ responses per dimension, following the same method used for LLM outputs. We then computed correlation between these aggregated human scores and those from three large LLMs: Mistral-Large, Qwen2.5-72B, and Llama3.1-70B. In addition, we calculated correlation between the original Likert-scale scores from SummEval and the checklist-based human scores. All correlations are statistically significant, indicating that CheckEval scores successfully capture human judgments.

\paragraph{Agreement} Due to the high annotation cost, we focused on \texttt{relevance} for agreement analysis. We collected binary annotations on 100 summaries (sample size selected based on a power analysis targeting 95\% confidence interval width of $\leq$ 0.2 for IEA scores). Each annotator answered approximately 10,000 questions. We report inter-annotator agreement among the three human annotators, as well as agreement between the human group and the large LLM group. We observe high agreement between humans as well as between humans and LLMs, showing that CheckEval elicits consistent scores across both human and LLM raters.

%% file: sections/7_analysis.tex
\begin{figure}[t]
    \centering    
    \includegraphics[width=0.9\columnwidth]{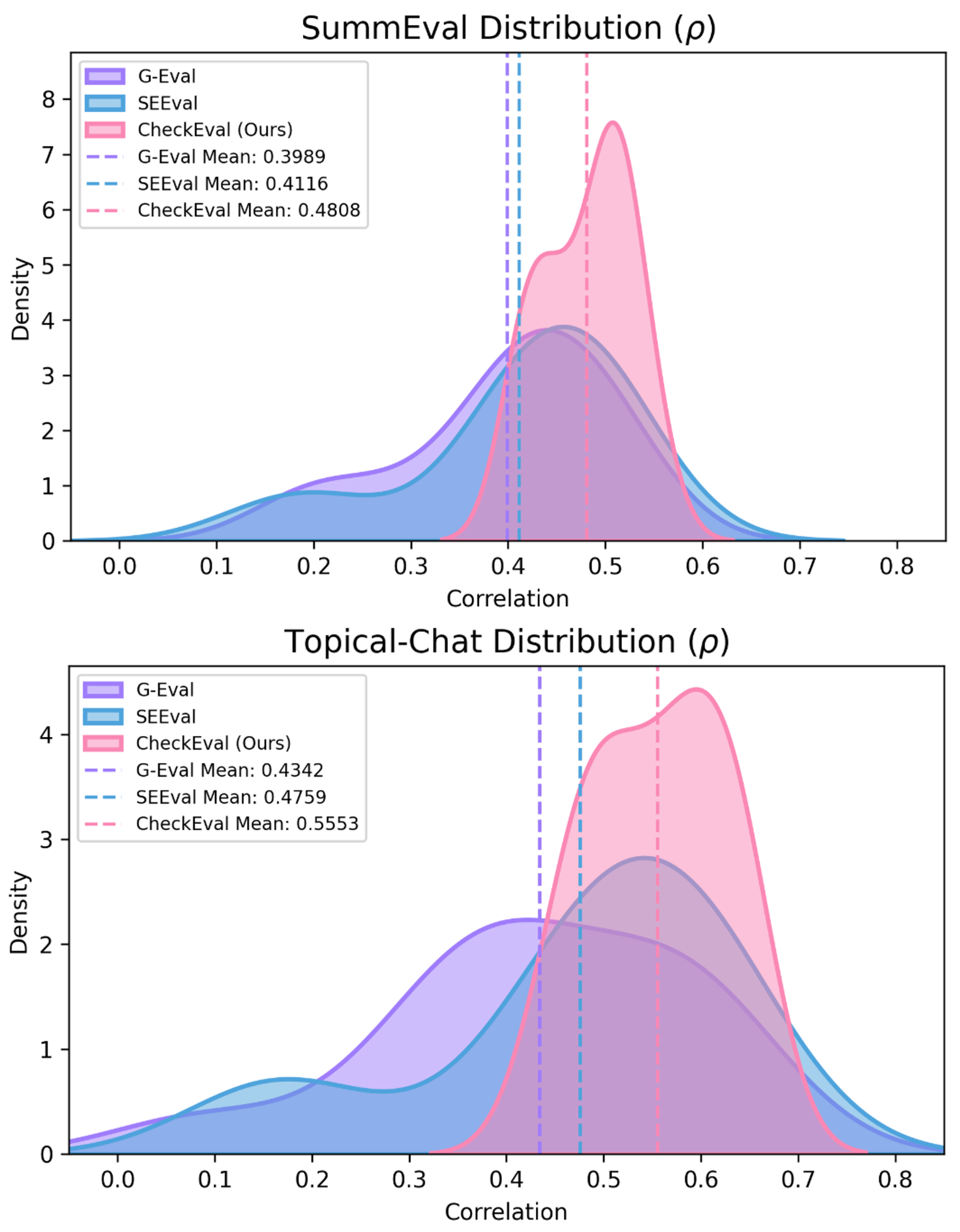}
    \caption{Kernel density estimation (KDE) of correlations with human judgments for G-Eval (purple), SEEval (blue) and CheckEval (pink) across different evaluator models on SummEval and Topical-Chat. Dashed lines indicate mean correlation values.}
    \label{fig:Dist-variance}
\end{figure}

\section{Analysis}
\subsection{Stability Analysis of Evaluation Methods}
We further analyze the stability of evaluation methods by examining the distribution of correlations with human judgments across different evaluator models. While the agreement metric (Section~\ref{subsec:agreement-analysis}) focuses on how consistently models assess the same samples, stability evaluates whether an evaluation method maintains reliable alignment with human annotations across all evaluator models. As shown in Figure~\ref{fig:Dist-variance}, CheckEval achieves higher mean correlation and lower variance than G-Eval and SEEval on both datasets, demonstrating more stable evaluation across different models. Detailed correlation statistics, including full mean and variance values, are available in Table~\ref{tab:mean-and-var}. 

\begin{table}[t]
\centering
\resizebox{\columnwidth}{!}{
    \begin{tabular}{@{}p{\textwidth}@{}}
    \toprule
    \textbf{Conversation history (source)} \\ \midrule
    \begin{tabularx}{\textwidth}{X}
    A: Hello, how are you today? Do you like to go to concerts? \\
    B: Not as much as I used to, but I do. \\
    A: Yeah, same here! Creed gave a concert so bad there were lawsuits against the band. \\
    ... \\
    B: I have no idea. I'm sure that someone has video of it. Do you enjoy the music of the Foo Fighters? \\
    A: Oh yes, I love them. I love the video of all the drummers and other instruments playing at the same time. People came from all over the world to be in that. \\
    B: They are pretty amazing. They performed a concert so loud that it showed up on New Zealand's seismic monitors! \\
    \end{tabularx} \\ \midrule
    
    \textbf{Context -} \textit{In 2002, a Creed concert was so bad that four concertgoers filed a lawsuit against the band.} \\ \midrule
    \textbf{System Response -} \textit{I know, I think I have heard that before, I think it was really cool.} \\ \midrule
    \textbf{Checklist} \\ \midrule
    \begin{tabularx}{\textwidth}{X c}
    \textbf{Questions} & \textbf{Answer} \\ \midrule
    Does the response avoid unnecessary repetition of the same content between sentences? & Yes \\
    Does each sentence directly relate to the topic being discussed? & No \\
    Is the overall message clear and easy to understand? & Yes \\
    Does each sentence in the response convey a clear meaning? & Yes \\
    Is the tone consistent throughout? & Yes \\
    Does the response avoid using jargon or overly complex words that might confuse the listener? & Yes \\
    Are there no major grammatical errors? & Yes \\
    Are there no ambiguous terms or phrases that could confuse the reader? & Yes \\
    \end{tabularx} \\ 
    \midrule
    \textbf{Raw Scores -} Human: \textbf{3} (1-3), G-Eval: \textbf{2} (1-5), CheckEval: \textbf{0.88} (0-1) \\ 
    \midrule
    \textbf{Normalized Scores  - } Human: \textbf{1} (0-1), G-Eval: \textbf{0.25} (0-1), CheckEval: \textbf{0.88} (0-1) \\ 
    \bottomrule
    \end{tabular}
    }
\caption{Case study on the \texttt{naturalness} dimension in the Topical-Chat.}
\label{tab:case_study}
\end{table}

\subsection{Case Study}
\label{subsec:case-study}
We conduct a case study on the \texttt{naturalness} dimension in the Topical-Chat dataset to illustrate how CheckEval enhances explainability by explicitly showing which evaluation criteria contribute to the final score (see Table~\ref{tab:case_study}). We evaluate system responses generated by Mistral-large, the model with the strongest correlation with human judgments. For this case study, we normalize all scores to a 0--1 scale for direct comparison. On evaluating the given text on naturalness, CheckEval (0.88) aligns more closely with human judgments (1.0), rating the response as natural. In contrast, G-Eval (0.25) assigned a much lower naturalness score. More importantly, while G-Eval provides only a score without explanation, CheckEval's systematic decomposition into specific sub-questions helps us attribute the high score to individual questions with a `yes’ answer (e.g., the response is natural because it avoids repetition, the message is clear, etc.). An additional case study on low-quality samples from benchmark datasets is presented in Appendix~\ref{sec:case-study-low}, further demonstrating how CheckEval operates across a wider range of text qualities.

\subsection{Ablation Study}
\label{subsec:ablation-study}
We conducted an ablation study to assess the contribution of each component in the CheckEval pipeline. Table~\ref{tab:checkeval-ablation} reports results when removing filtering and augmentation step. Both components contribute to overall performance, with the augmentation stage showing a slightly larger impact. We also explore whether the performance gap can be closed by increasing the baseline inference budget in Appendix~\ref{sec:inference budget}.

\begin{table}[t]
\centering
\setlength{\tabcolsep}{15pt}
\renewcommand{\arraystretch}{1.1}
\footnotesize
\resizebox{\columnwidth}{!}{%
\begin{tabular}{lcc}
\toprule
 & \textbf{SummEval} & \textbf{Topical-Chat} \\
\midrule
CheckEval & 0.48 & 0.55 \\
w/o filtering & 0.48 & 0.54 \\
w/o augmentation & 0.46 & 0.53 \\
\bottomrule
\end{tabular}}
\caption{Effect of filtering and augmentation components in CheckEval}
\label{tab:checkeval-ablation}
\end{table}

%% file: sections/8_conclusion.tex
\section{Conclusion}

We propose CheckEval, a fine-grained Boolean QA Checklist framework that addresses key limitations in existing LLM-as-Judge approaches for evaluating text generation. By decomposing evaluation criteria into structured binary questions, CheckEval enables reliable evaluation of (open-ended) text. Our experiments across various models and datasets demonstrate that CheckEval outperforms widely-adopted Likert scale-based methods like G-Eval, achieving higher correlation to human evaluation and IEA across different LLM evaluators. The framework shows particular strength in evaluating high-quality texts by effectively capturing subtle qualitative differences while maintaining explainability. Additionally, CheckEval enhances evaluation stability through reduced variance across LLMs. This shows that our framework offers a promising solution for constructing more reliable evaluation benchmarks across diverse NLG tasks.

%% file: sections/9_limitation.tex
\section{Limitations}

CheckEval improves the reliability of LLM-as-a-Judge evaluation, but it has several limitations. First, while automating checklist generation is a promising direction for improving scalability, it introduces challenges that are common to many automatic evaluation methods. CheckEval uses task-specific, human-written seed questions, which helps ground the evaluation in task-relevant criteria. However, as an automatic evaluation method, there may be factors beyond our control that lead to potential misalignment. In such cases, human involvement may be necessary to ensure alignment with task-specific goals. This is not a limitation of CheckEval specifically, but a broader challenge inherent to automatic evaluation approaches.

Second, this study focused on analyzing model-wise evaluation trends and comparing Likert-scale evaluation with Boolean QA checklist-based evaluation. However, recent LLM-as-a-Judge studies have introduced various techniques to enhance human alignment. Methods such as prompt optimization (e.g. chain-of-thought \cite{wei2022chain}, self-correction \cite{xu2023instructscore}), multi-agent debate \cite{chan2024chateval, kim-etal-2024-debate}, and meta-evaluator training \cite{kim2023prometheus, wu-etal-2024-instructeval, zhu2025judgelm} enable LLMs to make more enhanced judgments. Therefore, future work should compare it against these approaches and analyze how it differs in terms of reliability. This would also help determine whether CheckEval can be combined with such techniques to build a more robust evaluation framework.

Third, while CheckEval’s boolean-style decision improves evaluation reliability, not all NLG tasks and evaluation criteria can be strictly answered with a yes/no response. This limitation becomes more apparent when considering evaluation scenarios involving texts two to three times longer than those in the current benchmarks. As text length increases, some parts of a response may be strong while others are weak. For example, the first half of a response may be well-written and coherent, while the latter half is unclear or contains errors. This makes binary decisions insufficient for capturing subtle quality differences. The constraints of a yes/no format may become more pronounced in long-form evaluations, suggesting that future research should explore ways to mitigate this limitation while preserving the strengths of CheckEval. 

Fourth, CheckEval’s efficacy should be tested on a wider range of NLG tasks. While this study primarily focused on summarization and dialogue response generation, additional experiments are needed to validate CheckEval’s applicability to tasks such as story generation, long-form question answering, machine translation, and dialogue generation. Given that evaluation criteria vary by domain, it is important to examine how well CheckEval generalizes across different task settings. We note that generalizability of CheckEval is already actively being tested in follow-up work: for instance CheckEval has been used for tasks such as essay scoring \cite{10.1145/3706598.3713181}, creative writing evaluation \cite{lee2024navigating}, and healthcare evaluation \cite{mallinar2025scalable}. 

Finally, improving the automation of checklist design and evaluation processes would enhance CheckEval’s usability. Currently, checklist construction is a manual process tailored to specific tasks, making it difficult to predict the time and effort required for new evaluation domains. One potential solution is to pre-build a large-scale question database for NLG tasks and develop a system that automatically assembles relevant checklists based on task requirements. Future research should explore LLM-assisted checklist generation and reconfiguration methods to ensure that CheckEval can be efficiently applied to a broader range of tasks.

%% file: sections/10_acknowledgements.tex
\section{Acknowledgments}

We thank Meng-Chen Wu for his help during the rebuttal process. We also thank Jungsoo Park for discussions that helped shape the initial idea of CheckEval. Thanks to Soonwon Ka, Bokyung Son, and Keonwoo Kim for their useful feedback in the early stages of the project. We also appreciate the valuable discussion and support from Yulu Qin, tinlab at BU, Naver AX unsupervised learning and  DSBA NLP group at KU \& SNU. We acknowledge that the computational work reported in this paper was performed on the Shared Computing Cluster which is administered by \href{https://www.bu.edu/tech/support/research/}{Boston University's Research Computing Services}. In addition, JK and PK were supported by Institute of Information \& Communications Technology Planning \& Evaluation (IITP) grant funded by the Korea government (MSIT) (RS-2025-02214591, Development of an Innovative AI Agent for Worker-Friendly Autonomous Manufacturing). Also, JK and PK were supported by the BK21 FOUR Program (Education and Research Center for Industrial Innovation Analytics) funded by the Ministry of Education, Korea (No. 4120240214912)

%% file: sections/appendix.tex
\section{Detailed Experimental Setup}

\subsection{Baselines}
\label{subsec:baselines}

\paragraph{Baselines for main comparison (\Cref{tab:merged-results})}
(1) \textbf{BERTScore} \cite{zhang2019bertscore} calculates text similarity by contextual embeddings of BERT \cite{devlin2018bert}.
(2) \textbf{MoverScore} \cite{zhao2019moverscore} extends BERTScore by incorporating soft alignments, allowing words to be dynamically matched across texts. It refines similarity computation through an improved aggregation strategy that accounts for word importance and semantic shifts.
(3) \textbf{BARTScore} \cite{yuan2021bartscore} evaluates text quality by computing the average likelihood of a generated output under a BART-based conditional probability model. 
(4) \textbf{UniEval} \cite{zhong2022towards} is a multi-dimensional evaluation framework that assesses various dimensions of text generation by leveraging both reference-based and reference-free evaluation.
(5) \textbf{G-Eval} \cite{liu2023gpteval} is an LLM-based method, using chain-of-thought \cite{wei2022chain} and a form-filling paradigm to generate evaluation scores on a Likert scale. We select G-Eval as the main comparison point due to its widespread adoption \cite{liu2023x, liu-etal-2024-hd}, as well as considering the similarity between G-Eval and CheckEval that neither approach involves complex prompt engineering, additional model training or multi-agent evaluation.
(6) \textbf{SEEval} \cite{wu-etal-2025-seeval} is a prompt-based evaluator that incorporates self-explanation, guiding the model to justify its rating decisions without additional training.

\paragraph{Baselines for Comparative Analysis (\Cref{tab:evaluation-comparison})}
(1) \textbf{TIGERScore} \cite{jiang2024tigerscore} is a Llama 2 fine-tuned evaluation method that uses LLM to perform an explainable text evaluation.
(2) \textbf{GPTScore} \cite{fu2023gptscore} evaluates text by computing the conditional log-likelihood of reference or output generated under LLM.
(3) \textbf{Analyze-Rate} \cite{chiang-lee-2023-closer} analyzes how specific design choices in LLM-based evaluation, such as explanation prompting and output format, affect alignment with human judgment and finds that encouraging explanation improves correlation.
(4) \textbf{HD-EVAL} \cite{liu-etal-2024-hd} decomposes the evaluation into fine-grained criteria and trains a regression model to aggregate them in alignment with human preferences through iterative preference-based optimization.

\begin{table*}[t]
    \centering
    \setlength{\tabcolsep}{10pt}
    \renewcommand{\arraystretch}{1.3}
    \resizebox{0.95\textwidth}{!}{%
        \footnotesize
        \begin{tabular}{p{2cm} p{1.8cm} >{\centering\arraybackslash}p{0.7cm} >{\centering\arraybackslash}p{0.7cm} >{\centering\arraybackslash}p{0.7cm} >{\centering\arraybackslash}p{0.7cm} >{\centering\arraybackslash}p{0.7cm} >{\centering\arraybackslash}p{0.7cm} >{\centering\arraybackslash}p{0.7cm} >{\centering\arraybackslash}p{0.7cm} >{\centering\arraybackslash}p{0.7cm}}
            \toprule
            \multirow{2}{*}{\makecell[l]{\textbf{Model}\\ \textbf{}}} & \multirow{2}{*}{\makecell[l]{\textbf{Evaluation}\\ \textbf{Methods}}} & \multicolumn{3}{c}{\textbf{CNN}} & \multicolumn{3}{c}{\textbf{Xsum}} & \multicolumn{3}{c}{\textbf{Average}} \\
            \cmidrule(lr){3-5} \cmidrule(lr){6-8} \cmidrule(lr){9-11}
            & & \textbf{$r$} & \textbf{$\rho$} & \textbf{$\tau$} & \textbf{$r$} & \textbf{$\rho$} & \textbf{$\tau$} & \textbf{$r$} & \textbf{$\rho$} & \textbf{$\tau$} \\
            \midrule
            \cellcolor{pastelPink}{\textbf{Llama3.1-70B}}
            & G-Eval & 0.5097 & 0.4559 & 0.4261 & 0.2317 & 0.2317 & 0.2317 & 0.3707 & 0.3438 & 0.3289 \\
            \cellcolor{pastelPink} & CheckEval & 0.7002 & 0.6747 & 0.5683 & 0.5394 & 0.5018 & 0.4355 & \textbf{0.6198} & \textbf{0.5883} & \textbf{0.5019} \\
            \cdashline{2-11}
            \cellcolor{pastelPink}{\textbf{Mistral-Large}}
            & G-Eval & 0.5617 & 0.6104 & 0.5705 & 0.5834 & 0.5834 & 0.5834 & 0.5726 & 0.5969 & 0.5770 \\
            \cellcolor{pastelPink} & CheckEval & 0.7472 & 0.7291 & 0.6277 & 0.5889 & 0.5825 & 0.5352 & \textbf{0.6681} & \textbf{0.6558} & \textbf{0.5815} \\
            \cdashline{2-11}
            \cellcolor{pastelPink}{\textbf{Qwen2.5-72B}}
            & G-Eval & 0.6830 & 0.7154 & 0.6686 & 0.5236 & 0.5236 & 0.5236 & 0.6033 & \textbf{0.6195} & \textbf{0.5961} \\
            \cellcolor{pastelPink} & CheckEval & 0.7312 & 0.7013 & 0.6078 & 0.4931 & 0.4898 & 0.4197 & \textbf{0.6122} & 0.5956 & 0.5138 \\
            \midrule
            \cellcolor{pastelBlue}{\textbf{Mistral-Small}}
            & G-Eval & 0.5656 & 0.5425 & 0.5070 & 0.4833 & 0.4833 & 0.4833 & 0.5245 & 0.5129 & \textbf{0.4952} \\
            \cellcolor{pastelBlue} & CheckEval & 0.6563 & 0.6211 & 0.5239 & 0.4950 & 0.4496 & 0.3890 & \textbf{0.5757} & \textbf{0.5354} & 0.4565 \\
            \cdashline{2-11}
            \cellcolor{pastelBlue}{\textbf{Gemma2-27B}}
            & G-Eval & 0.6124 & 0.6543 & 0.6115 & 0.5644 & 0.5644 & 0.5644 & \textbf{0.5884} & \textbf{0.6094} & \textbf{0.5880} \\
            \cellcolor{pastelBlue} & CheckEval & 0.6975 & 0.6493 & 0.5397 & 0.4547 & 0.4040 & 0.3482 & 0.5761 & 0.5267 & 0.4440 \\
            \cdashline{2-11}
            \cellcolor{pastelBlue}{\textbf{Qwen2.5-32B}}
            & G-Eval & 0.6487 & 0.6357 & 0.5941 & 0.4290 & 0.4290 & 0.4290 & 0.5389 & 0.5324 & 0.5116 \\
            \cellcolor{pastelBlue} & CheckEval & 0.7286 & 0.7132 & 0.6145 & 0.5532 & 0.5231 & 0.4547 & \textbf{0.6409} & \textbf{0.6182} & \textbf{0.5346} \\
            \midrule
            \cellcolor{pastelOlive}{\textbf{Llama3.1-8B}}
            & G-Eval & 0.2785 & 0.2228 & 0.2082 & 0.0614 & 0.0614 & 0.0614 & 0.1700 & 0.1421 & 0.1348 \\
            \cellcolor{pastelOlive} & CheckEval & 0.6100 & 0.5995 & 0.4924 & 0.4244 & 0.4292 & 0.3669 & \textbf{0.5172} & \textbf{0.5144} & \textbf{0.4297} \\
            \cdashline{2-11}
            \cellcolor{pastelOlive}{\textbf{Gemma2-9B}}
            & G-Eval & 0.6599 & 0.7002 & 0.6544 & 0.5546 & 0.5546 & 0.5546 & \textbf{0.6073} & \textbf{0.6274} & \textbf{0.6045} \\
            \cellcolor{pastelOlive} & CheckEval & 0.5353 & 0.5713 & 0.4597 & 0.4502 & 0.4529 & 0.3875 & 0.4928 & 0.5121 & 0.4236 \\
            \cdashline{2-11}
            \cellcolor{pastelOlive}{\textbf{Qwen2.5-7B}}
            & G-Eval & 0.4688 & 0.4307 & 0.4025 & 0.2137 & 0.2137 & 0.2137 & 0.3413 & 0.3222 & 0.3081 \\
            \cellcolor{pastelOlive} & CheckEval & 0.6157 & 0.5672 & 0.4775 & 0.4419 & 0.4681 & 0.4063 & \textbf{0.5288} & \textbf{0.5177} & \textbf{0.4419} \\
            \midrule
            \cellcolor{pastelLavender}{\textbf{GPT-4 Turbo}}
            & G-Eval & 0.4941 & 0.5402 & 0.5049 & 0.5560 & 0.5560 & 0.5560 & 0.5251 & 0.5481 & 0.5305 \\
            \cellcolor{pastelLavender} & CheckEval & 0.7155 & 0.7211 & 0.6363 & 0.5922 & 0.5658 & 0.4961 & \textbf{0.6539} & \textbf{0.6435} & \textbf{0.5662} \\
            \cdashline{2-11}
            \cellcolor{pastelLavender}{\textbf{GPT-4o}}
            & G-Eval & 0.2864 & 0.3100 & 0.2897 & 0.0582 & 0.0582 & 0.0582 & 0.1723 & 0.1841 & 0.1740 \\
            \cellcolor{pastelLavender} & CheckEval & 0.6724 & 0.6601 & 0.5452 & 0.5448 & 0.5282 & 0.4564 & \textbf{0.6086} & \textbf{0.5942} & \textbf{0.5008} \\
            \cdashline{2-11}
            \cellcolor{pastelLavender}{\textbf{GPT-4o-mini}}
            & G-Eval & 0.5424 & 0.5833 & 0.5136 & 0.4591 & 0.4591 & 0.4212 & 0.5008 & 0.5212 & 0.4674 \\
            \cellcolor{pastelLavender} & CheckEval & 0.6175 & 0.6340 & 0.5451 & 0.4394 & 0.4831 & 0.4591 & \textbf{0.5285} & \textbf{0.5586} & \textbf{0.5021} \\
            \bottomrule
        \end{tabular}
    }
    \caption{Average correlation scores across dimensions on the QAGS-CNN and QAGS-Xsum. we report $r$, $\rho$ and $\tau$. Colors indicate model groups: large (pink), medium (blue), small (green) and  GPT (purple).}
    \label{tab:qags-correlation} 
\end{table*}

\subsection{Detailed Process of Seed Question Writing}
We constructed seed questions based on predefined evaluation criteria (e.g., coherence, consistency), aiming for atomic, conceptually clear, and non-overlapping formulations. Each evaluation dimension was first decomposed into finer-grained sub-dimensions, and a set of seed questions was written to cover each sub-dimension. This ensured both conceptual coverage and balance across dimensions. To guide this process, we consulted prior task-specific literature (e.g., summarization evaluation papers) and followed established guidelines where available. We observed that overly fine-grained seed questions often led LLMs to generate augmented variants that deviated from the original intent. Therefore, we intentionally maintained an appropriate granularity level to preserve alignment throughout augmentation.
All seed questions were cross-validated by our team to ensure clarity, consistency, and relevance across different evaluation dimensions.

\subsection{Top-3 Models per Evaluation Method}
\label{subsec:top3models}

The following models achieved the highest correlation with human judgments for each evaluation method: 
\textbf{CheckEval} (SummEval: \texttt{GPT-4-Turbo}, \texttt{Mistral-Large}, \texttt{Gemma2-27B}; Topical-Chat: \texttt{GPT-4-Turbo}, \texttt{GPT-4o}, \texttt{Mistral-Large}), \textbf{G-Eval} (SummEval: \texttt{GPT-4-Turbo}, \texttt{GPT-4o-mini}, \texttt{Mistral-Large}; Topical-Chat: \texttt{GPT-4-Turbo}, \texttt{Mistral-Large}, \texttt{Qwen2.5-72B}), and \textbf{SEEval} (SummEval: \texttt{Mistral-Large}, \texttt{GPT-4-Turbo}, \texttt{Qwen2.5-32B}; Topical-Chat: \texttt{Mistral-Large}, \texttt{Qwen2.5-72B}, \texttt{GPT-4-Turbo}).

\subsection{Human Validation}
\label{sec:humaneval-setting}

To validate CheckEval, we conducted three human evaluation studies (correlation, agreement study:~\Cref{subsec:human-validation} and Checklist Validation~\Cref{fig:checklist-validation}). For these studies, summaries were randomly sampled from the SummEval dataset using stratification based on original human annotation scores to ensure balanced coverage across quality levels. Each study involves three Ph.D student-level evaluators. We recruited three human evaluators with Ph.D. student-level qualifications or above in Computer Science, all of whom had a background in evaluation research and summarization/dialogue tasks. Each participant was compensated with a gift card equivalent to approximately 10,000 KRW ($\approx$ 7 USD) per hour.\footnote{Note that the annotation was conducted in South Korea, where the compensation level is slightly above the local minimum wage.}

For the correlation study (Table~\ref{tab:humaneval} - Correlation), 20 summaries are randomly sampled from the SummEval dataset. These summaries are subsequently evaluated on a binary (yes/no) basis against a checklist comprising four dimensions: \texttt{coherence}, \texttt{consistency}, \texttt{fluency}, and \texttt{relevance}.

For the agreement study (Table~\ref{tab:humaneval} - Agreement), 100 summaries are sampled from the SummEval dataset. These summaries are then evaluated on a binary (yes/no) basis concerning only \texttt{relevance} due to practical cost constraints (evaluation this dimension alone already requires each annotator to answer approximately 10K questions). The sample size of 100 was calculated from a power analysis based on a pilot study.

For the checklist validation study (Figure~\ref{fig:checklist-validation}), each annotator saw the same set of items, with approximately 28 questions per evaluation dimension in SummEval and 26 in Topical-Chat.

\begin{table}[t]
\centering
\setlength{\tabcolsep}{5pt}
\renewcommand{\arraystretch}{1.3}
\resizebox{0.95\columnwidth}{!}{%
\footnotesize
    \begin{tabular}{llcccc}
    \toprule
    \multirow{2}{*}{\makecell[l]{\textbf{Model}\\ \textbf{Group}}}  
    & \multirow{2}{*}{\makecell[l]{\textbf{Evaluation}\\ \textbf{Methods}}}  
    & \multicolumn{2}{c}{\textbf{CNN}}  
    & \multicolumn{2}{c}{\textbf{Xsum}} \\ 
    \cmidrule(lr){3-4} \cmidrule(lr){5-6}
    & & \textbf{$\alpha$} & \textbf{$\kappa$}  
    & \textbf{$\alpha$} & \textbf{$\kappa$} \\
    \midrule
    \multirow{2}{*}{All} & G-Eval & 0.2215 & 0.3624 & 0.2873 & 0.2853 \\
    & CheckEval & \textbf{0.4149} & \textbf{0.4149} & \textbf{0.3416} & \textbf{0.3416} \\
    \midrule
    \multirow{2}{*}{Large} & G-Eval & 0.1595 & 0.3345 & 0.1166 & 0.3772 \\
    & CheckEval & \textbf{0.6420} & \textbf{0.6420} & \textbf{0.5189} & \textbf{0.5189} \\
    \midrule
    \multirow{2}{*}{Medium} & G-Eval & 0.0526 & 0.5612 & 0.0546 & 0.3458 \\
    & CheckEval & \textbf{0.5971} & \textbf{0.5970} & \textbf{0.4074} & \textbf{0.4074} \\
    \midrule
    \multirow{2}{*}{Small} & G-Eval & 0.0805 & 0.0761 & 0.1796 & 0.0440 \\
    & CheckEval & \textbf{0.0846} & \textbf{0.0846} & \textbf{0.1881} & \textbf{0.1880} \\
    \midrule
    \multirow{2}{*}{GPT} & G-Eval & 0.0625 & 0.3920 & 0.1674 & 0.2156 \\
    & CheckEval & \textbf{0.4720} & \textbf{0.4719} & \textbf{0.2998} & \textbf{0.2997} \\
    \midrule
    \multirow{2}{*}{Top-3} & G-Eval & 0.0489 & 0.4845 & 0.0349 & 0.4381 \\
    & CheckEval & \textbf{0.5234} & \textbf{0.5234} & \textbf{0.5066} & \textbf{0.5066} \\
    \bottomrule
    \end{tabular}}
\caption{IEA - QAGS}
\label{tab:iea-qags}
\end{table}

\section{Additional Results}

\subsection{Additional experiments with QAGS}
\label{sec:appendix-qags-experiments}

Table \ref{tab:qags-correlation} shows the correlation between various evaluation methods and human judgments on the QAGS dataset. The results show that CheckEval outperforms G-Eval for 9 out of the 12 LLMs (comparable to results on the other two datasets reported in the main text), indicating its effectiveness as an evaluation Framework. Furthermore, Table \ref{tab:iea-qags} compares the IEA of G-Eval and CheckEval on the QAGS dataset. Across all model groups, CheckEval consistently achieves a higher IEA than G-Eval, demonstrating its advantage in robustness.

\subsection{Comparative performance of various LLM-as-a-Judge methods}

We also included a broader comparison with recent evaluation methods surveyed in \citet{gu2024survey, Gao2024LLMbasedNE}. For CheckEval and G-Eval, we use scores using the best-performing evaluator in our experiments (Mistral-large). Table~\ref{tab:evaluation-comparison} shows that CheckEval performs well overall on both datasets, and remains competitive even compared to more recent approaches. However, we would like to emphasize again that our main goal is not to propose the best-performing LLM-as-a-judge method. Instead, our focus is on building a more reliable evaluation process and analyzing its consistency across different LLMs, and that is why comparison to G-Eval is the most directly relevant result.

\subsection{Checklist Validation}

To quantify the reliability of human annotations in the checklist validation study, we adopted Exact Match as our IAA metric over more common alternatives like Fleiss’s Kappa. This choice was motivated by two characteristics of our data. The evaluation results showed a response distribution heavily concentrated on `Yes' (or 1) due to the high quality of the items (see Figure~\Cref{fig:checklist-validation}), which can make Kappa's chance correction misleading. Furthermore, the small number of items per dimension (fewer than 30) can impact the stability of Kappa scores. Given these factors, we report Exact Match scores of 0.677 for SummEval and 0.634 for Topical-Chat (see Table~\ref{tab:checklist-em}).

\begin{table}[t]
\centering
\setlength{\tabcolsep}{5pt}
\renewcommand{\arraystretch}{1.4}
\resizebox{\columnwidth}{!}{%
\footnotesize
\begin{tabular}{lccccc}
\toprule
\textbf{SummEval} & Coh. & Con. & Flu. & Rel. & Avg. \\
\cmidrule(l){2-6}
EM                & 0.7330    & 0.6920      & 0.7100  & 0.5710    & 0.6765 \\
\midrule
\textbf{Topical-Chat} & Coh. & Eng. & Gro. & Nat. & Avg. \\
\cmidrule(l){2-6}
EM                    & 0.6470    & 0.6670       & 0.6300       & 0.5910     & 0.6338 \\
\bottomrule
\end{tabular}}
\caption{Agreement (Exact Match) for each dimension in checklist validation.}
\label{tab:checklist-em}
\end{table}

\begin{table}[t]
\centering
\setlength{\tabcolsep}{2pt}
\footnotesize
\renewcommand{\arraystretch}{1.1}
\resizebox{\columnwidth}{!}{%
\begin{tabular}{llcccc}
\toprule
\multirow{2}{*}{\makecell[l]{\textbf{Evaluation}\\ \textbf{Methods}}} & \textbf{Model} & \multicolumn{2}{c}{\textbf{SummEval (Avg.)}} & \multicolumn{2}{c}{\textbf{Topical-Chat (Avg.)}} \\
& & $\rho$ & $\tau$ & $\rho$ & $r$ \\
\midrule
TIGERScore       & LLaMA 2–13B$^\dagger$         & 0.39 & 0.31 & 0.28 & 0.26 \\
GPTScore         & GPT-4               & 0.39 & 0.34 & 0.36 & 0.34 \\
G-Eval           & Mistral-large       & 0.52 & 0.47 & 0.64 & 0.62 \\
Analyze-Rate     & Claude 3 Sonnet     & 0.53 & 0.44 & 0.64 & 0.64 \\
HD-EVAL          & GPT-4               & 0.53 & –    & 0.62 & 0.63 \\
SEEval          & Claude 3 Sonnet     & 0.52 & 0.47 & 0.65 & 0.64 \\
\textbf{CheckEval} & \textbf{Mistral-large} & \textbf{0.55} & \textbf{0.48} & \textbf{0.65} & \textbf{0.65} \\
\bottomrule
\end{tabular}}
\caption{Comparative performance of various LLM-as-a-Judge methods. Models marked with $\dagger$ are fine-tuned.}
\label{tab:evaluation-comparison}
\end{table}

\section{Discussion}
\subsection{Analysis of Performance on High and Low-Quality Texts}
\label{sec:case-study-low}

As LLMs improve, their high-quality outputs become more fluent and coherent, making it increasingly difficult for evaluation methods to differentiate subtle quality differences. 
Meanwhile, low-quality text poses a different challenge, as its overall readability is low, obscuring distinctions between evaluation criteria and making it harder to properly assess all target dimensions of quality. 
Given these differences, it is important to assess how evaluation methods handle varying levels of text quality. 
To this end, we conduct a detailed dimension-wise analysis by dividing the data into high-quality and low-quality groups based on human annotation scores (e.g., on a 1--5 scale, treat scores $\geq$3 as High, $<$3 as Low). We compute the average correlation across 12 LLMs to analyze how CheckEval and G-Eval align with human judgments for different levels of text quality. 

\begin{figure}[t]
   \centering
   \begin{subfigure}{\columnwidth}
       \includegraphics[width=0.95\columnwidth]{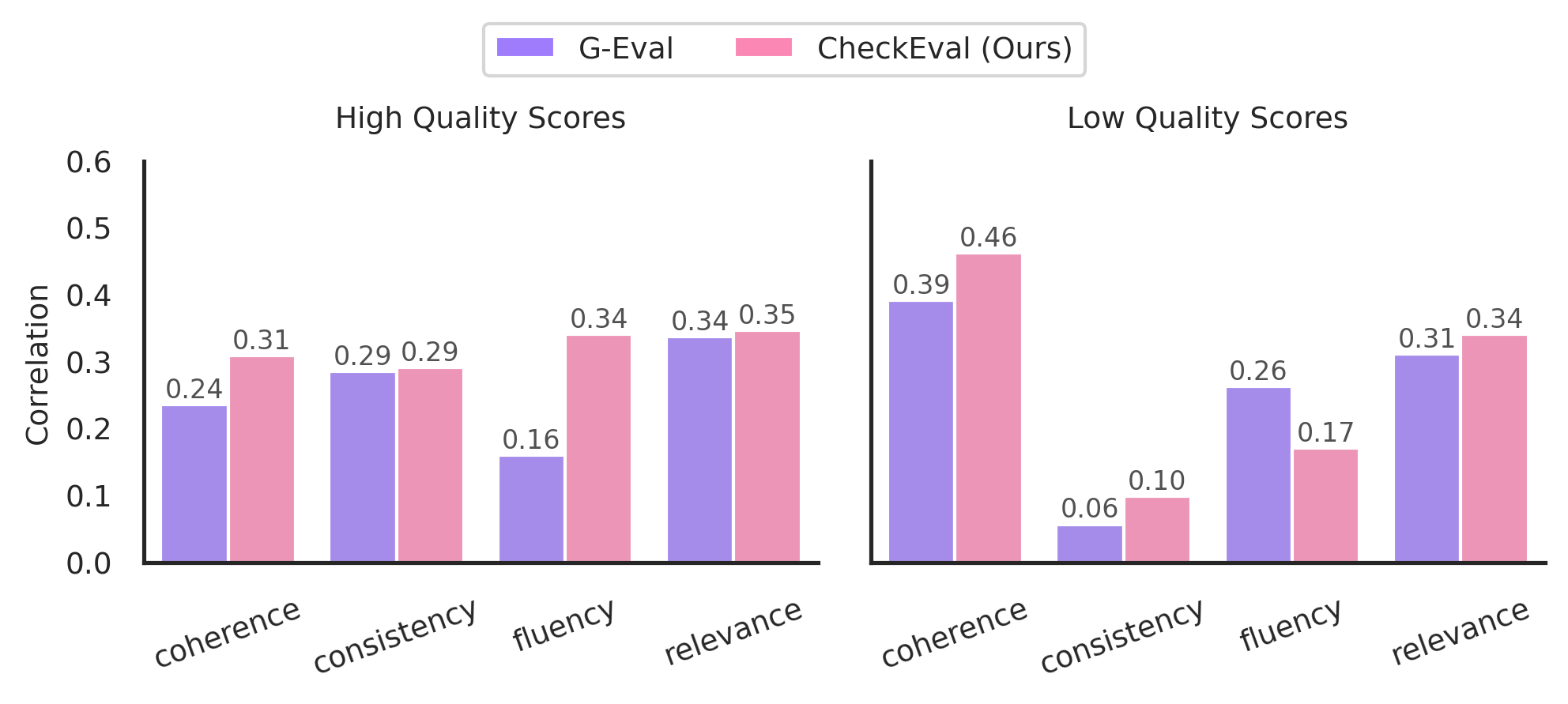}
       \caption{SummEval}
       \label{fig:Qual - Summeval}
   \end{subfigure}
   \begin{subfigure}{\columnwidth}
       \includegraphics[width=0.95\columnwidth]{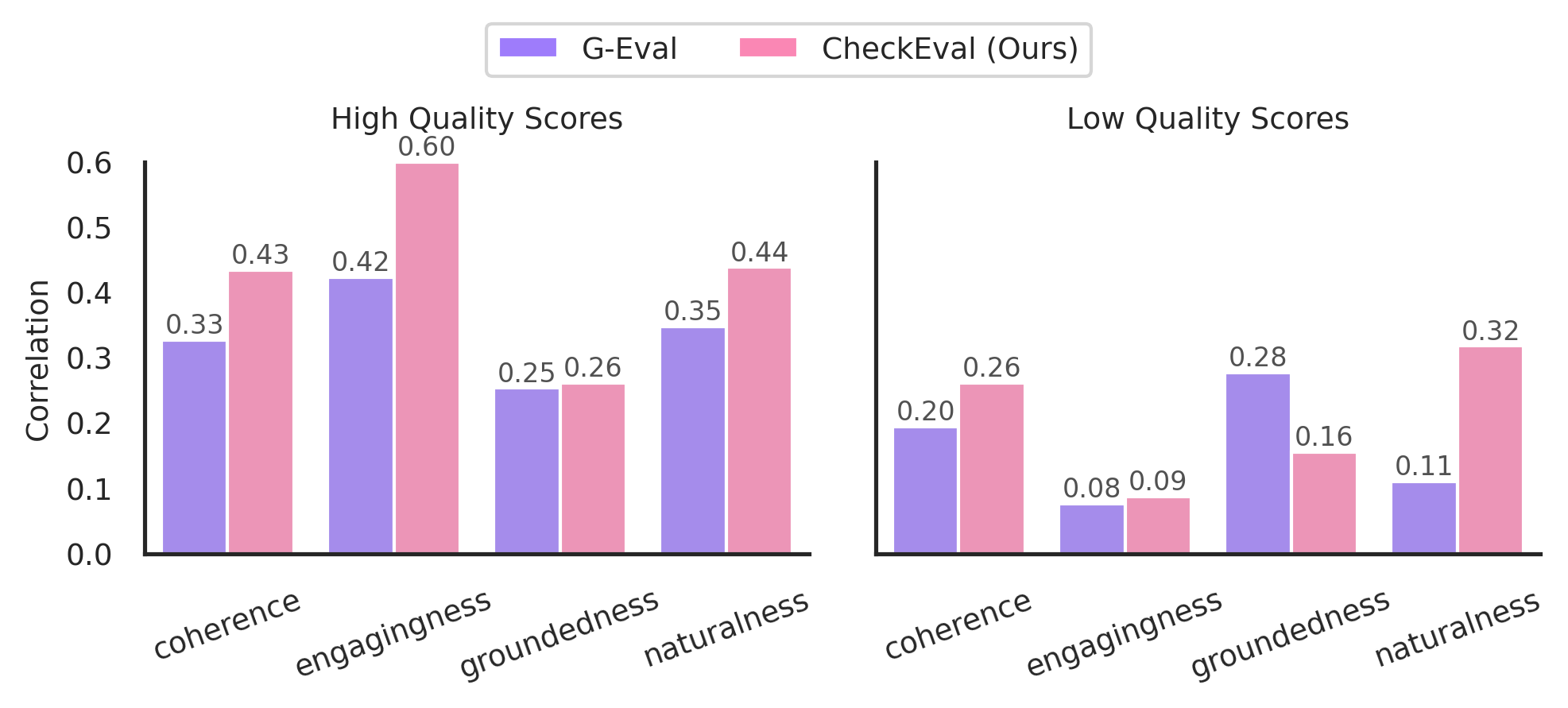}
       \caption{Topical-Chat}
       \label{fig:Qual - Topical}
   \end{subfigure}
   \caption{dimension-wise correlation analysis of G-Eval (purple) and CheckEval (pink), with samples divided based on human annotator ratings into High-Quality (human ratings $\geq$3) and Low-Quality (human ratings $<$3) groups. Each bar represents correlation with human judgments across different quality dimensions.}
   \label{fig:Quality-analysis}
\end{figure}

As shown in Figure~\ref{fig:Quality-analysis}, CheckEval consistently achieves higher correlations with human judgments than G-Eval in high-quality texts across all dimensions. Notably, for SummEval, CheckEval shows much stronger alignment in \texttt{fluency} (0.34 vs. 0.16). For Topical-Chat, it outperforms G-Eval in \texttt{engagingness} (0.60 vs. 0.42) and \texttt{naturalness} (0.44 vs. 0.35) by a large margin. 

However, for low-quality texts, while CheckEval generally maintains stronger correlations compared to G-Eval, it exhibits performance drops in a small number of cases, notably in \texttt{fluency} (SummEval) and \texttt{groundedness} (Topical-Chat). From our additional analysis of the results, one possible explanation is that discrepancies between benchmark definitions and actual human annotations of these dimensions may have contributed to the observed performance drop in CheckEval. For example, while SummEval defines \texttt{fluency} as the absence of formatting issues, capitalization errors, or ungrammatical sentence structures that hinder readability, human annotators often prioritized overall readability over strict grammatical correctness. Since CheckEval relies on fine-grained Boolean QA decisions aligned with predefined criteria, the correlation with human scores may be impacted when human annotation practices deviate from the exact evaluation guidelines. In the \texttt{groundedness} dimension of Topical-Chat, a different issue arises. For low-quality texts, CheckEval’s strict yes/no framework often results in uniformly low scores, making it difficult to distinguish between varying degrees of poor responses. In contrast, G-Eval, which allows for more gradient judgments, showed advantages in these cases. This suggests potential refinements to the Boolean QA framework to better handle annotation inconsistencies while preserving its fine-grained evaluation capability.

\begin{table}[t]
    \centering
    \renewcommand{\arraystretch}{1.2}
    \resizebox{\columnwidth}{!}{%
        \footnotesize
        \begin{tabular}{p{2cm} p{1.7cm} >{\centering\arraybackslash}p{2.3cm} >{\centering\arraybackslash}p{2.3cm}}
            \toprule
            \multirow{2}{*}{\makecell[l]{\textbf{Model}\\ \textbf{}}} & \multirow{2}{*}{\makecell[l]{\textbf{Aggregation}\\ \textbf{Strategy}}} & \multicolumn{2}{c}{\textbf{SummEval (Avg.) }}\\
            \cmidrule(l){3-4}
            & & \textbf{$\rho$} & \textbf{$\tau$} \\
            \midrule
            \cellcolor{pastelPink}\textbf{Llama3.1-70B}         & original & 0.4628 & 0.4037 \\
            \cellcolor{pastelPink}         & weighted & 0.4674 ($\pm$0.015) & 0.4046 ($\pm$0.016) \\
            \cdashline{2-4}
            \cellcolor{pastelPink}\textbf{Mistral-Large}         & original & 0.5486 & 0.4797 \\
            \cellcolor{pastelPink}         & weighted & 0.5320 ($\pm$0.021) & 0.4622 ($\pm$0.021) \\
            \cdashline{2-4}
            \cellcolor{pastelPink}\textbf{Qwen2.5-72B}           & original & 0.5024 & 0.4413 \\
            \cellcolor{pastelPink}           & weighted & 0.5002 ($\pm$0.0130) & 0.4360 ($\pm$0.006) \\
            \midrule
            \cellcolor{pastelBlue}\textbf{Mistral-Small}         & original & 0.4473 & 0.3938 \\
            \cellcolor{pastelBlue}         & weighted & 0.4424 ($\pm$0.029) & 0.3920 ($\pm$0.029) \\
            \cdashline{2-4}
            \cellcolor{pastelBlue}\textbf{Gemma2-27B}            & original & 0.5108 & 0.4426 \\
            \cellcolor{pastelBlue}           & weighted & 0.5063 ($\pm$0.008) & 0.4361 ($\pm$0.006) \\
            \cdashline{2-4}
            \cellcolor{pastelBlue}\textbf{Qwen2.5-32B}           & original & 0.5193 & 0.4566 \\
            \cellcolor{pastelBlue}           & weighted & 0.5093 ($\pm$0.006) & 0.4422 ($\pm$0.005) \\
            \midrule
            \cellcolor{pastelOlive}\textbf{Llama3.1-8B}         & original & 0.4342 & 0.3654 \\
            \cellcolor{pastelOlive}         & weighted & 0.3752 ($\pm$0.009) & 0.3191 ($\pm$0.008) \\
            \cdashline{2-4}
            \cellcolor{pastelOlive}\textbf{Gemma2-9B}            & original & 0.4186 & 0.3607 \\
            \cellcolor{pastelOlive}            & weighted & 0.4561 ($\pm$0.005) & 0.3920 ($\pm$0.003) \\
            \cdashline{2-4}
            \cellcolor{pastelOlive}\textbf{Qwen2.5-7B}           & original & 0.4162 & 0.3652 \\
            \cellcolor{pastelOlive}           & weighted & 0.4026 ($\pm$0.023) & 0.3545 ($\pm$0.019) \\
            \midrule
            \cellcolor{pastelLavender}\textbf{GPT-4 Turbo}       & original & 0.5212 & 0.4633 \\
            \cellcolor{pastelLavender}       & weighted & 0.5182 ($\pm$0.003) & 0.4563 ($\pm$0.001) \\
            \cdashline{2-4}
            \cellcolor{pastelLavender}\textbf{GPT-4o}            & original & 0.5042 & 0.4377 \\
            \cellcolor{pastelLavender}            & weighted & 0.4771 ($\pm$0.026) & 0.4113 ($\pm$0.023) \\
            \cdashline{2-4}
            \cellcolor{pastelLavender}\textbf{GPT-4o-mini}       & original & 0.4913 & 0.4157 \\
            \cellcolor{pastelLavender}       & weighted & 0.4817 ($\pm$0.013) & 0.4032 ($\pm$0.008) \\
            \bottomrule
        \end{tabular}
    }
    \caption{Effect of question weighting strategy on SummEval.}
    \label{tab:weighted_checkeval}
\end{table}

\subsection{Does CheckEval need question weighting?}
\label{sec:weighted question aggregation}

We conducted an additional analysis to investigate whether incorporating question-specific weights improves the reliability of CheckEval scores. Motivated by HD-Eval \cite{liu-etal-2024-hd}, we trained a linear regression model using 20\% of the SummEval data to estimate the relative importance (i.e., weights) of each checklist question. These weights were then used to compute a weighted CheckEval score. To assess robustness, the process was repeated across five random seeds, each sampling a different 20\% subset of the data. Table~\ref{tab:weighted_checkeval} reports the average results and standard deviation across seeds. ``original'' denotes the unweighted CheckEval score, while ``weighted'' denotes the score after applying the learned question-specific weights. The overall results were mixed. A couple of evaluator models benefited from learning the weights, but most others did not. Since there were no reliable gains from weighting the questions, we ultimately chose not to incorporate weighted aggregation into our results. While we only experimented with a simple linear weighting strategy here, we could explore more sophisticated methods of estimating question importance as well as learning weights that are generalizable across different evaluator models in future work.

\begin{table}[t]
\centering
\setlength{\tabcolsep}{7pt}
\footnotesize
\renewcommand{\arraystretch}{1.3}
\resizebox{\columnwidth}{!}{%
\begin{tabular}{lllll}
\toprule
\textbf{Dataset} & \textbf{Dimension} & \textbf{Original} & \textbf{Binary} & \textbf{Binary} \\
\textbf{} & \textbf{} & \textbf{Scale} & \textbf{Conversion 1} & \textbf{Conversion 2} \\
\midrule
SummEval & All & 1–5 & [4,5] $\rightarrow$ 1 & [3,4,5] $\rightarrow$ 1\\
 &  &  & [1,2,3] $\rightarrow$ 0 & [1,2] $\rightarrow$ 0 \\
Topical Chat & Coh./Eng./Nat. & 1–3 & [3] $\rightarrow$ 1& [2,3] $\rightarrow$ 1\\
 &  &  & [1,2] $\rightarrow$ 0 & [1] $\rightarrow$ 0 \\
Topical Chat & Gro. & 0–1 & – & – \\
QAGS & All & 0–1 & – & – \\
\bottomrule
\end{tabular}}
\caption{Binary conversion schemes applied to G-Eval’s Likert-scale outputs to enable fairer IEA comparison.}
\label{tab:binary-mapping}
\end{table}

\subsection{Does Binarizing Likert-Scale Outputs Close the IEA Gap?}
\label{binarizing-likert}

We conducted an additional analysis to investigate whether the observed IEA gap is a fundamental difference between the evaluation protocols or simply an artifact of their different output formats (binary vs. Likert). One way to test this would be to directly binarize the Likert scores derived from the evaluator models. We conducted this experiment with G-Eval’s Likert-scale outputs--that is, we converted the Likert scores (1-5 scale for SummEval and 1-3 scale for Topical-Chat) to binary (0/1) scores by mapping the lower values to 0 and higher values to 1. To ensure that the results are not affected by the mapping choice of the middle value on the scale, we tested both possible versions of the mapping schemes: treating the middle value as 0 and 1, respectively, as detailed in Table~\ref{tab:binary-mapping}. Scores of evaluation dimensions that already employed a binary scoring scheme were not converted.

The results, shown in Tables~\ref{tab:binary-summ} and~\ref{tab:binary-topical}, are clear and consistent. While binarizing the outputs does improve G-Eval’s IEA scores compared to using the original Likert scale scores, a large performance gap to CheckEval remains across all model groups. We therefore conclude that the performance difference is not solely an effect of the output format but stems from the fundamental improvements in our proposed checklist-based evaluation protocol.

\begin{table}[t]
\centering
\footnotesize
\setlength{\tabcolsep}{7pt}
\renewcommand{\arraystretch}{1.3}
\resizebox{0.95\columnwidth}{!}{%
\begin{tabular}{llcc}
\toprule
\textbf{Model Size} & \textbf{Method} & \textbf{$\alpha$} & \textbf{$\kappa$} \\
\midrule
\textbf{All} & G-Eval & 0.0929 & 0.1859 \\
 & G-Eval (binary [4,5]→1) & 0.1063 & 0.2812 \\
 & G-Eval (binary [3,4,5]→1) & 0.1074 & 0.2835 \\
 & CheckEval & \textbf{0.4803} & \textbf{0.4803} \\
\midrule
\textbf{Best} & G-Eval & 0.0731 & 0.2266 \\
 & G-Eval (binary [4,5]→1) & 0.0666 & 0.4650 \\
 & G-Eval (binary [3,4,5]→1) & 0.0666 & 0.4647 \\
 & CheckEval & \textbf{0.6471} & \textbf{0.6471} \\
\midrule
\textbf{GPT} & G-Eval & 0.0841 & 0.2018 \\
 & G-Eval (binary [4,5]→1) & 0.0693 & 0.3012 \\
 & G-Eval (binary [3,4,5]→1) & 0.0676 & 0.3016 \\
 & CheckEval & \textbf{0.5575} & \textbf{0.5575} \\
\midrule
\textbf{Large} & G-Eval & 0.0512 & 0.1586 \\
 & G-Eval (binary [4,5]→1) & 0.3204 & 0.4646 \\
 & G-Eval (binary [3,4,5]→1) & 0.3228 & 0.4575 \\
 & CheckEval & \textbf{0.6731} & \textbf{0.6731} \\
\midrule
\textbf{Medium} & G-Eval & 0.0430 & 0.1411 \\
 & G-Eval (binary [4,5]→1) & 0.0606 & 0.2758 \\
 & G-Eval (binary [3,4,5]→1) & 0.0658 & 0.2821 \\
 & CheckEval & \textbf{0.5617} & \textbf{0.5617} \\
\midrule
\textbf{Small} & G-Eval & 0.0635 & 0.0998 \\
 & G-Eval (binary [4,5]→1) & 0.1450 & 0.1984 \\
 & G-Eval (binary [3,4,5]→1) & 0.0835 & 0.1995 \\
 & CheckEval & \textbf{0.2387} & \textbf{0.2387} \\
\bottomrule
\end{tabular}}
\caption{IEA on SummEval after converting G-Eval’s Likert-scale outputs to binary formats.}
\label{tab:binary-summ}
\end{table}

\begin{table}[t]
\centering
\footnotesize
\renewcommand{\arraystretch}{1.3}
\resizebox{0.9\columnwidth}{!}{%
\begin{tabular}{llcc}
\toprule
\textbf{Model Size} & \textbf{Method} & \textbf{$\alpha$} & \textbf{$\kappa$} \\
\midrule
\textbf{All} & G-Eval & 0.0589 & 0.3407 \\
 & G-Eval (binary [3]→1) & 0.0565 & 0.3841 \\
 & G-Eval (binary [2,3]→1) & 0.1711 & 0.3893 \\
 & CheckEval & \textbf{0.4494} & \textbf{0.4494} \\
\midrule
\textbf{Best} & G-Eval & 0.0255 & 0.5593 \\
 & G-Eval (binary [3]→1) & 0.0181 & 0.5799 \\
 & G-Eval (binary [2,3]→1) & 0.0181 & 0.5806 \\
 & CheckEval & \textbf{0.5736} & \textbf{0.5736} \\
\midrule
\textbf{GPT} & G-Eval & 0.0385 & 0.4971 \\
 & G-Eval (binary [3]→1) & 0.0231 & 0.5196 \\
 & G-Eval (binary [2,3]→1) & 0.0240 & 0.5151 \\
 & CheckEval & \textbf{0.5395} & \textbf{0.5395} \\
\midrule
\textbf{Large} & G-Eval & 0.0145 & 0.5088 \\
 & G-Eval (binary [3]→1) & 0.0090 & 0.5749 \\
 & G-Eval (binary [2,3]→1) & 0.0092 & 0.5753 \\
 & CheckEval & \textbf{0.6736} & \textbf{0.6736} \\
\midrule
\textbf{Medium} & G-Eval & 0.0688 & 0.2231 \\
 & G-Eval (binary [3]→1) & 0.0585 & 0.2450 \\
 & G-Eval (binary [2,3]→1) & 0.0613 & 0.2410 \\
 & CheckEval & \textbf{0.5044} & \textbf{0.5043} \\
\midrule
\textbf{Small} & G-Eval & 0.0372 & 0.1636 \\
 & G-Eval (binary [3]→1) & 0.0635 & 0.1713 \\
 & G-Eval (binary [2,3]→1) & 0.0674 & 0.1591 \\
 & CheckEval & \textbf{0.1669} & \textbf{0.1668} \\
\bottomrule
\end{tabular}}
\caption{IEA on Topical-Chat after converting G-Eval’s Likert-scale outputs to binary formats.}
\label{tab:binary-topical}
\end{table}

\subsection{Does Increasing the Inference Budget Strengthen the Baselines?} 
\label{sec:inference budget}

To address the possibility that our performance gains stem from differences in the inference budget, we increased the budget for the baseline. One straightforward way to do this is to sample multiple outputs and aggregate the results. We applied this method to G-Eval on Topical-Chat, setting `temperature=1.0' to enable diverse generations and using `n=3' samples before averaging the scores. As shown in Table~\ref{tab:inference_budget}, the resulting correlations changed minimally ($r$ 0.6387 vs. 0.6389; $\rho$ 0.6169 vs. 0.6176), indicating that this aggregation does not close the performance gap with our checklist-based approach.

\begin{table}[h!]
\centering
\setlength{\tabcolsep}{8pt}
\renewcommand{\arraystretch}{1.1}
\resizebox{\columnwidth}{!}{%
\begin{tabular}{lllrr}
\toprule
 Dataset         & Correlation &    Method    &   Mean &  Variance \\
\midrule
\multirow{1}{*}{SummEval} 
                 & \multirow{1}{*}{Spearman} 
                              & G-Eval       & 0.3989 &    0.0100 \\
                 &            & SEEval       & 0.4116 &    0.0129 \\
                 &            & CheckEval    & 0.4808 &    0.0019 \\
                 & \multirow{1}{*}{Kendall}  
                              & G-Eval       & 0.3647 &    0.0084 \\
                 &            & SEEval       & 0.3684 &    0.0111 \\
                 &            & CheckEval    & 0.4163 &    0.0016 \\
\midrule
\multirow{1}{*}{Topical-Chat} 
                 & \multirow{1}{*}{Spearman} 
                              & G-Eval       & 0.4342 &    0.0220 \\
                 &            & SEEval       & 0.4759 &    0.0245 \\
                 &            & CheckEval    & 0.5553 &    0.0043 \\
                 & \multirow{1}{*}{Pearson}  
                              & G-Eval       & 0.4797 &    0.0205 \\
                 &            & SEEval       & 0.4679 &    0.0231 \\
                 &            & CheckEval    & 0.5546 &    0.0042 \\
\bottomrule
\end{tabular}}
\caption{Mean and variance for each dataset and correlation method}
\label{tab:mean-and-var}
\end{table}

\begin{table}[h!]
\centering
\setlength{\tabcolsep}{8pt}
\renewcommand{\arraystretch}{1.1}
\resizebox{\columnwidth}{!}{%
\begin{tabular}{llcc}
\toprule
\textbf{Aspect} & \textbf{Metric} & $\rho$ & $r$  \\
\midrule
\multirow{4}{*}{Mistral-Large} 
  & G-Eval & 0.6389 & 0.6176 \\
  & G-Eval aggregation (n=3) & 0.6387 & 0.6169 \\
  & SEEVal & 0.6352 & 0.6323 \\
  & CheckEval & 0.6451 & 0.6453 \\
\bottomrule
\end{tabular}}
\caption{Comparison of Evaluation Methods under Different Inference Budgets.}
\label{tab:inference_budget}
\end{table}

\section{The number of questions at each stage}
\label{sec:appendix-number of questions}
We provide a step-by-step breakdown of the number of questions, from the initial seed questions through the augmentation and filtering stages to the final checklist, with the number of questions varying across different dimensions. Before and after filtering, the correlation shows slight variations. For the SummEval, Spearman’s $\rho$ changed from 0.4790 to 0.4816, while Kendall’s $\tau$ changed from 0.4143 to 0.4163. In the Topical-Chat, Pearson’s $r$ remained unchanged at 0.5553, whereas Spearman’s $\rho$ increased from 0.5446 to 0.5546. The number of questions for each dataset is reported in Table~\ref{tab:checklist-statistics-summeval} and ~\ref{tab:checklist-statistics-topicalchat}, respectively.

\section{Open-source model information}
\label{sec:appendix-link}

Table~\ref{tab:model-link} provides links to all open-source models used in our experiments. Table~\ref{tab:models-license} lists each model along with its corresponding license. Table~\ref{tab:datasets-license} summarizes the datasets used and their associated licenses. If a dataset is publicly available but no explicit license is provided, we denote the license as `–' in the table.

\begin{table}[h!]
\centering
\begin{adjustbox}{max width=\linewidth}
\begin{tabular}{lcccc}
\toprule
\textbf{} & \textbf{Coherence} & \textbf{Consistency} & \textbf{Fluency} & \textbf{Relevance} \\ 
\midrule
Seed Questions & 3 & 3 & 4 & 5 \\ 
Diversification & 7 & 12 & 11 & 5 \\ 
Elaboration & 13 & 14 & 24 & 21 \\ 
Filtered Questions & 0 & 0 & 4 & 5 \\ 
Final Checklist & 23 & 29 & 35 & 26 \\ 
\bottomrule
\end{tabular}
\end{adjustbox}
\caption{The number of questions - SummEval.}
\label{tab:checklist-statistics-summeval}
\end{table}

\begin{table}[ht]
\centering
\begin{adjustbox}{max width=\linewidth}
\begin{tabular}{lcccc}
\toprule
\textbf{} & \textbf{Naturalness} & \textbf{Coherence} & \textbf{Engagingness} & \textbf{Groundedness} \\ 
\midrule
Seed Questions & 5 & 4 & 4 & 5 \\ 
Diversification & 9 & 6 & 10 & 6 \\ 
Elaboration & 14 & 11 & 17 & 15 \\ 
Filtered Questions & 0 & 1 & 0 & 0 \\ 
Final Checklist & 28 & 20 & 31 & 26 \\ 
\bottomrule
\end{tabular}
\end{adjustbox}
\caption{The number of questions - Topical-Chat.}
\label{tab:checklist-statistics-topicalchat}
\end{table}

\begin{table}[h!]
\centering
\begin{adjustbox}{max width=\columnwidth}
\begin{tabular}{ll}
\toprule
\textbf{Model} & \textbf{Link} \\ \midrule
Llama3.1-70B & \href{https://huggingface.co/meta-llama/Llama-3.1-70B-Instruct}{https://huggingface.co/meta-llama/Llama-3.1-70B-Instruct} \\ 
Mistral-large (123B) & \href{https://huggingface.co/mistralai/Mistral-Large-Instruct-2411}{https://huggingface.co/mistralai/Mistral-Large-Instruct-2411} \\ 
Qwen2.5-72B & \href{https://huggingface.co/Qwen/Qwen2.5-72B-Instruct}{https://huggingface.co/Qwen/Qwen2.5-72B-Instruct} \\ 
Mistral-Small (22B) & \href{https://huggingface.co/mistralai/Mistral-Small-Instruct-2409}{https://huggingface.co/mistralai/Mistral-Small-Instruct-2409} \\ 
Gemma2-27B & \href{https://huggingface.co/google/gemma-2-27b}{https://huggingface.co/google/gemma-2-27b-it} \\ 
Qwen2.5-32B & \href{https://huggingface.co/Qwen/Qwen2.5-32B-Instruct}{https://huggingface.co/Qwen/Qwen2.5-32B-Instruct} \\ 
Llama3.1-8B & \href{https://huggingface.co/meta-llama/Llama-3.1-8B-Instruct}{https://huggingface.co/meta-llama/Llama-3.1-8B-Instruct} \\ 
Gemma2-9B & \href{https://huggingface.co/google/gemma-2-9b-it}{https://huggingface.co/google/gemma-2-9b-it} \\ 
Qwen2.5-7B & \href{https://huggingface.co/Qwen/Qwen2.5-7B-Instruct}{https://huggingface.co/Qwen/Qwen2.5-7B-Instruct} \\ 
\bottomrule
\end{tabular}
\end{adjustbox}
\caption{Model Links.}
\label{tab:model-link}
\end{table}

\begin{table}[h!]
    \centering
    \setlength{\tabcolsep}{12pt}
    \renewcommand{\arraystretch}{1.1}
    \normalsize
    \resizebox{\columnwidth}{!}{%
        \begin{tabular}{ll}
            \toprule
            \textbf{Models} & \textbf{License} \\
            \midrule
            \texttt{meta-llama/Llama-3.1-70B-Instruct} & llama3.1 \\
            \texttt{mistralai/Mistral-Large-Instruct-2411} & mrl \\
            \texttt{Qwen/Qwen2.5-72B-Instruct} & qwen \\
            \texttt{mistralai/Mistral-Small-Instruct-2409} & mrl \\
            \texttt{google/gemma-2-27b-it} & gemma \\
            \texttt{Qwen/Qwen2.5-32B-Instruct} & Apache license 2.0 \\
            \texttt{meta-llama/Llama-3.1-8B-Instruct} & llama3.1 \\
            \texttt{google/gemma-2-9b-it} & gemma \\
            \texttt{Qwen/Qwen2.5-7B-Instruct} & Apache license 2.0 \\
            \texttt{GPT-4 Turbo} & Proprietary \\
            \texttt{GPT-4o} & Proprietary \\
            \texttt{GPT-4o-mini} & Proprietary \\
            \bottomrule
        \end{tabular}
    }
    \caption{List of models and their corresponding licenses.}
    \label{tab:models-license}
\end{table}

\begin{table}[h!]
    \centering
    \setlength{\tabcolsep}{30pt}
    \renewcommand{\arraystretch}{1.1}
    \normalsize
    \resizebox{0.9\columnwidth}{!}{%
        \begin{tabular}{ll}
            \toprule
            \textbf{Datasets} & \textbf{License} \\
            \midrule
            SummEval & MIT license \\
            Topical-chat & CDLA-Sharing-1.0 \\
            QAGS & -\\
            \bottomrule
        \end{tabular}
    }
    \caption{List of datasets and their corresponding licenses.}
    \label{tab:datasets-license}
\end{table}

\section{Prompts}
\label{sec:appendix-prompt}
Figure~\ref{fig:evaluation_prompt_summeval} and \ref{fig:evaluation_prompt_topicalchat} shows the detailed evaluation prompt. Figure~\ref{fig:augmentation_prompt_diversification} and \ref{fig:augmentation_prompt_elaboration} shows the detailed augmentation prompt. Figure~\ref{fig:filtering_prompt} shows the filtering prompt.

\begin{table*}[t]
\centering
\setlength{\tabcolsep}{9pt}
\renewcommand{\arraystretch}{1.2}
\resizebox{\textwidth}{!}{%
\footnotesize
\begin{tabular}{p{2cm}p{1.8cm}cccccccc!{\vrule}cc}
\toprule
\multirow{2}{*}{\makecell[l]{\textbf{Model}}} 
& \multirow{2}{*}{\makecell[l]{\textbf{Evaluation}\\ \textbf{Method}}} 
& \multicolumn{2}{c}{\textbf{Coherence}} 
& \multicolumn{2}{c}{\textbf{Consistency}} 
& \multicolumn{2}{c}{\textbf{Fluency}} 
& \multicolumn{2}{c}{\textbf{Relevance}} 
& \multicolumn{2}{c}{\textbf{Average}} \\
& & $\rho$ & $\tau$ & $\rho$ & $\tau$ & $\rho$ & $\tau$ & $\rho$ & $\tau$ & $\rho$ & $\tau$ \\
\midrule
\multicolumn{12}{l}{\textbf{\textit{LLM-as-a-judge}}} \\
\midrule
\cellcolor{pastelPink}\textbf{Llama3.1-70B} & G-Eval & 0.5206 & 0.4459 & 0.3513 & 0.3306 & 0.3104 & 0.2924 & 0.4371 & 0.3800 & 0.4048 & 0.3622 \\
\cellcolor{pastelPink} & SEEval & 0.5836 & 0.4821 & 0.4188 & 0.3878 & 0.2287 & 0.2043 & 0.4037 & 0.3295 & 0.4087 & 0.3509 \\
\cellcolor{pastelPink} & CheckEval & 0.6222 & 0.5264 & 0.5406 & 0.4913 & 0.2637 & 0.2288 & 0.4248 & 0.3682 & \textbf{0.4628} & \textbf{0.4037} \\
\cdashline{2-12}
\cellcolor{pastelPink}\textbf{Mistral-Large} & G-Eval & 0.5892 & 0.5078 & 0.6153 & 0.5824 & 0.3611 & 0.3435 & 0.5026 & 0.4368 & 0.5171 & 0.4676 \\
\cellcolor{pastelPink} & SEEval & 0.5472 & 0.5132 & 0.6065 & 0.5782 & 0.4563 & 0.4352 & 0.5406 & 0.4581 & 0.5377 & 0.4962 \\
\cellcolor{pastelPink} & CheckEval & 0.6439 & 0.5424 & 0.6132 & 0.5668 & 0.4563 & 0.3926 & 0.4811 & 0.4169 & \textbf{\phantom{0}0.5486$^\ast$} & \textbf{\phantom{0} 0.4797$^\ast$ } \\
\cdashline{2-12}
\cellcolor{pastelPink}\textbf{Qwen2.5-72B} & G-Eval & 0.3937 & 0.3420 & 0.5248 & 0.4903 & 0.3202 & 0.3050 & 0.4762 & 0.4178 & 0.4287 & 0.3888 \\
\cellcolor{pastelPink} & SEEval & 0.4761 & 0.4002 & 0.5156 & 0.4742 & 0.3746 & 0.3452 & 0.5118 & 0.4390 & 0.4695 & 0.4147 \\
\cellcolor{pastelPink} & CheckEval & 0.5778 & 0.4932 & 0.5490 & 0.5047 & 0.4113 & 0.3582 & 0.4717 & 0.4092 & \textbf{0.5025} & \textbf{0.4413} \\
\midrule
\cellcolor{pastelBlue}\textbf{Mistral-Small} & G-Eval & 0.2885 & 0.2463 & 0.2748 & 0.2532 & 0.0134 & 0.0126 & 0.1629 & 0.1343 & 0.1849 & 0.1616 \\
\cellcolor{pastelBlue} & SEEval & 0.1260 & 0.1003 & 0.2040 & 0.1823 & 0.3822 & 0.3519 & 0.1829 & 0.1468 & 0.2238 & 0.1953 \\
\cellcolor{pastelBlue} & CheckEval & 0.5297 & 0.4531 & 0.5113 & 0.4712 & 0.3098 & 0.2670 & 0.4381 & 0.3837 & \textbf{0.4472} & \textbf{0.3937} \\
\cdashline{2-12}
\cellcolor{pastelBlue}\textbf{Gemma2-27B} & G-Eval & 0.5731 & 0.4951 & 0.5111 & 0.4684 & 0.1596 & 0.1520 & 0.5239 & 0.4515 & 0.4419 & 0.3917 \\
\cellcolor{pastelBlue} & SEEval & 0.5892 & 0.5021 & 0.4829 & 0.4552 & 0.3629 & 0.2132 & 0.5193 & 0.4361 & 0.4886 & 0.4017 \\
\cellcolor{pastelBlue} & CheckEval & 0.6199 & 0.5244 & 0.4924 & 0.4485 & 0.4402 & 0.3756 & 0.4906 & 0.4220 & \textbf{0.5108} & \textbf{0.4426} \\
\cdashline{2-12}
\cellcolor{pastelBlue}\textbf{Qwen2.5-32B} & G-Eval & 0.5361 & 0.4682 & 0.5550 & 0.5199 & 0.3606 & 0.3420 & 0.5363 & 0.4703 & 0.4970 & \textbf{0.4501} \\
\cellcolor{pastelBlue} & SEEval & 0.5731 & 0.4681 & 0.5578 & 0.5267 & 0.3893 & 0.3460 & 0.4352 & 0.4371 & 0.4889 & 0.4445 \\
\cellcolor{pastelBlue} & CheckEval & 0.6056 & 0.4938 & 0.5311 & 0.4767 & 0.4879 & 0.4157 & 0.4605 & 0.3797 & \textbf{0.5213} & 0.4415 \\
\midrule
\cellcolor{pastelOlive}\textbf{Llama3.1-8B} & G-Eval & 0.2689 & 0.2253 & 0.2988 & 0.2763 & 0.0088 & 0.0087 & 0.3644 & 0.3139 & 0.2352 & 0.2060 \\
\cellcolor{pastelOlive} & SEEval & 0.2684 & 0.2190 & 0.0508 & 0.0483 & 0.1623 & 0.1472 & 0.1488 & 0.1251 & 0.1576 & 0.1349 \\
\cellcolor{pastelOlive} & CheckEval & 0.5045 & 0.4048 & 0.4561 & 0.3887 & 0.3040 & 0.2654 & 0.3933 & 0.3168 & \textbf{0.4145} & \textbf{0.3439} \\
\cdashline{2-12}
\cellcolor{pastelOlive}\textbf{Gemma2-9B} & G-Eval & 0.5649 & 0.4895 & 0.4555 & 0.4206 & -0.0252 & -0.0221 & 0.5272 & 0.4602 & 0.3806 & 0.3370 \\
\cellcolor{pastelOlive} & SEEval & 0.5636 & 0.4843 & 0.4045 & 0.3935 & 0.2876 & 0.2510 & 0.4520 & 0.4548 & 0.4269 & 0.3959 \\
\cellcolor{pastelOlive} & CheckEval & 0.5777 & 0.4876 & 0.3979 & 0.3450 & 0.2798 & 0.2358 & 0.4590 & 0.4003 & \textbf{0.4286} & \textbf{0.3672} \\
\cdashline{2-12}
\cellcolor{pastelOlive}\textbf{Qwen2.5-7B} & G-Eval & 0.3785 & 0.3270 & 0.5343 & 0.5020 & 0.3309 & 0.3146 & 0.4154 & 0.3617 & 0.4148 & \textbf{0.3763} \\
\cellcolor{pastelOlive} & SEEval & 0.3950 & 0.3259 & 0.4767 & 0.4373 & 0.2595 & 0.2352 & 0.4350 & 0.3623 & 0.3916 & 0.3402 \\
\cellcolor{pastelOlive} & CheckEval & 0.4068 & 0.3398 & 0.4214 & 0.3800 & 0.4598 & 0.4226 & 0.3768 & 0.3183 & \textbf{0.4162} & 0.3652 \\
\midrule
\cellcolor{pastelLavender}\textbf{GPT-4 Turbo} & G-Eval & 0.4912 & 0.4251 & 0.6498 & 0.6229 & 0.3878 & 0.3668 & 0.5064 & 0.4397 & 0.5088 & \textbf{0.4636} \\
\cellcolor{pastelLavender} & SEEval & 0.5292 & 0.4621 & 0.6351 & 0.6031 & 0.3551 & 0.3327 & 0.4728 & 0.4501 & 0.4981 & 0.4620 \\
\cellcolor{pastelLavender} & CheckEval & 0.5807 & 0.4901 & 0.6232 & 0.5872 & 0.4611 & 0.4058 & 0.4197 & 0.3713 & \textbf{0.5212} & \textbf{0.4636} \\
\cdashline{2-12}
\cellcolor{pastelLavender}\textbf{GPT-4o} & G-Eval & 0.1896 & 0.1581 & 0.4219 & 0.3911 & 0.2862 & 0.2676 & 0.3969 & 0.3421 & 0.3237 & 0.2897 \\
\cellcolor{pastelLavender} & SEEval & 0.3391 & 0.3618 & 0.4421 & 0.4162 & 0.3665 & 0.3512 & 0.4021 & 0.3617 & 0.3875 & 0.3727 \\
\cellcolor{pastelLavender} & CheckEval & 0.5564 & 0.4644 & 0.5304 & 0.4738 & 0.4699 & 0.4125 & 0.4602 & 0.4001 & \textbf{0.5042} & \textbf{0.4377} \\
\cdashline{2-12}
\cellcolor{pastelLavender}\textbf{GPT-4o-mini} & G-Eval & 0.4826 & 0.4197 & 0.5243 & 0.4837 & 0.2734 & 0.2598 & 0.5192 & 0.4524 & 0.4499 & 0.4039 \\
\cellcolor{pastelLavender} & SEEval & 0.5149 & 0.4221 & 0.4831 & 0.4567 & 0.3552 & 0.3005 & 0.4882 & 0.4681 & 0.4604 & 0.4119 \\
\cellcolor{pastelLavender} & CheckEval & 0.5854 & 0.4829 & 0.4939 & 0.4286 & 0.3883 & 0.3314 & 0.4975 & 0.4199 & \textbf{0.4913} & \textbf{0.4157} \\
\bottomrule
\end{tabular}
}
\caption{Sample-level Spearman ($\rho$) and Kendall tau ($\tau$) correlations on the SummEval.  The best score per model category is \textbf{bolded}, and the highest overall score is marked with \textbf{*}.}
\label{tab:main-corr-result-summeval}
\end{table*}

\begin{table*}
    \centering
    \setlength{\tabcolsep}{10pt} 
    \renewcommand{\arraystretch}{1.2}
    \resizebox{\textwidth}{!}{%
\label{tab:iea}
\begin{tabular}{l l ll ll ll ll ll} 
\toprule
\multirow{2}{*}{
\begin{tabular}[c]{@{}l@{}}\textbf{Model}\\\textbf{Group}\end{tabular}
} & 
\multirow{2}{*}{
\begin{tabular}[c]{@{}l@{}}\textbf{Evaluation}\\\textbf{Methods}\end{tabular}
} & 
\multicolumn{2}{c}{\textbf{Coherence}} & 
\multicolumn{2}{c}{\textbf{Consistency}} & 
\multicolumn{2}{c}{\textbf{Fluency}} & 
\multicolumn{2}{c}{\textbf{Relevance}} & 
\multicolumn{2}{c}{\textbf{Average}} \\
& & 
\multicolumn{1}{c}{$\alpha$} & 
\multicolumn{1}{c}{$\kappa$} & 
\multicolumn{1}{c}{$\alpha$} & 
\multicolumn{1}{c}{$\kappa$} & 
\multicolumn{1}{c}{$\alpha$} & 
\multicolumn{1}{c}{$\kappa$} & 
\multicolumn{1}{c}{$\alpha$} & 
\multicolumn{1}{c}{$\kappa$} & 
\multicolumn{1}{c}{$\alpha$} & 
\multicolumn{1}{c}{$\kappa$} \\
\midrule
All & G-Eval & 0.0751 & 0.2706 & 0.0539 & 0.1625 & 0.1626 & 0.0699 & 0.0799 & 0.2407 & 0.0929 & 0.1859\\
    & SEEval & 0.0713 & 0.1332 & 0.0837 & 0.1457 & 0.0789 & 0.1391 & 0.0861 & 0.1420 & 0.0800 & 0.1400 \\
    & CheckEval & 0.4242 & 0.4242 & 0.2963 & 0.2963 & 0.4422 & 0.4422 & 0.7584 & 0.7584 & 0.4803 & 0.4803 \\
\midrule
Large & G-Eval & 0.0448 & 0.2170 & 0.0476 & 0.0057 & 0.0621 & 0.2372 & 0.0502 & 0.1745 & 0.0512 & 0.1586\\
    & SEEval & 0.0531 & 0.1827 & 0.0674 & 0.1965 & 0.0592 & 0.1884 & 0.0603 & 0.1924 & 0.0600 & 0.1900 \\
    & CheckEval & 0.7154 & 0.7154 & 0.5757 & 0.5757 & 0.5207 & 0.5206 & 0.8806 & 0.8806 & 0.6731 & 0.6731\\
\midrule
Medium & G-Eval & 0.0096 & 0.3742 & 0.0229 & 0.1306 & 0.0970 & -0.1462 & 0.0424 & 0.2057 & 0.0430 & 0.1411\\
    & SEEval & 0.0826 & 0.1234 & 0.0947 & 0.1361 & 0.0883 & 0.1292 & 0.0944 & 0.1313 & 0.0900 & 0.1300 \\
    & CheckEval & 0.6455 & 0.6455 & 0.2723 & 0.2723 & 0.5851 & 0.5851 & 0.7440 & 0.7440 & 0.5617 & 0.5617\\
\midrule
Small & G-Eval & 0.0704 & 0.2237 & 0.0044 & 0.1351 & 0.1089 & -0.1161 & 0.0702 & 0.1564 & 0.0635 & 0.0998\\
    & SEEval & 0.0117 & 0.0628 & 0.0265 & 0.0741 & 0.0189 & 0.0663 & 0.0229 & 0.0768 & 0.0200 & 0.0700 \\
    & CheckEval & 0.0827 & 0.0826 & 0.0237 & 0.0237 & 0.1746 & 0.1746 & 0.6739 & 0.6739 & 0.2387 & 0.2387\\
\midrule
GPT & G-Eval & 0.1425 & 0.1513 & 0.0984 & 0.0823 & 0.0064 & 0.3388 & 0.0889 & 0.2347 & 0.0841 & 0.2018\\
    & SEEval & 0.1196 & 0.3097 & 0.1338 & 0.3289 & 0.1275 & 0.3158 & 0.1391 & 0.3256 & 0.1300 & 0.3200 \\
    & CheckEval & 0.5081 & 0.5081 & 0.4135 & 0.4135 & 0.5473 & 0.5473 & 0.7612 & 0.7612 & 0.5575 & 0.5575\\
\midrule
Top-3 & G-Eval & 0.1104 & 0.2360 & 0.1002 & 0.0544 & 0.0171 & 0.3751 & 0.0647 & 0.2407 & 0.0731 & 0.2266\\
    & SEEval & 0.0812 & 0.1786 & 0.0973 & 0.1962 & 0.0884 & 0.1927 & 0.0931 & 0.1925 & 0.0900 & 0.1900 \\
    & CheckEval & 0.6236 & 0.6236 & 0.4836 & 0.4836 & 0.6698 & 0.6698 & 0.8114 & 0.8114 & 0.6471 & 0.6471\\
\bottomrule
\end{tabular}
    }
    \caption{IEA - SummEval.}
    \label{tab:iea-full result-SummEval}
\end{table*}

\begin{table*}[ht]
    \centering
    \setlength{\tabcolsep}{9.5pt}
    \renewcommand{\arraystretch}{1.2}
    \resizebox{\textwidth}{!}{%
        \footnotesize
        \begin{tabular}{p{2cm}p{1.8cm}cccccccc!{\vrule}cc}
            \toprule
            \multirow{2}{*}{\makecell[l]{\textbf{Model}}}  
            & \multirow{2}{*}{\makecell[l]{\textbf{Evaluation}\\ \textbf{Methods}}} 
            & \multicolumn{2}{c}{\textbf{Coherence}} 
            & \multicolumn{2}{c}{\textbf{Engagingness}} 
            & \multicolumn{2}{c}{\textbf{Groundedness}} 
            & \multicolumn{2}{c}{\textbf{Naturalness}} 
            & \multicolumn{2}{c}{\textbf{Average}} \\
            & & $\rho$ & $r$ & $\rho$ & $r$ & $\rho$ & $r$ & $\rho$ & $r$ & $\rho$ & $r$ \\
            \midrule
            \multicolumn{12}{l}{\textbf{\textit{LLM-as-a-judge}}} \\
            \midrule
            \cellcolor{pastelPink}\textbf{Llama3.1-70B} & G-Eval
                & 0.4089 & 0.3622 
                & 0.3968 & 0.3501 
                & 0.6190 & 0.5553 
                & 0.3684 & 0.2991 
                & 0.4483 & 0.3917 \\
            \cellcolor{pastelPink} & SEEval
                & 0.5160 & 0.4923
                & 0.6384 & 0.6312
                & 0.6091 & 0.6164
                & 0.4223 & 0.4238
                & 0.5465 & 0.5409 \\
            \cellcolor{pastelPink} & CheckEval
                & 0.5517 & 0.5360 
                & 0.6547 & 0.6551 
                & 0.4706 & 0.4917 
                & 0.6065 & 0.6082 
                & \textbf{0.5709} & \textbf{0.5727} \\
            \cdashline{2-12}
            \cellcolor{pastelPink}\textbf{Mistral-Large} & G-Eval
                & 0.5709 & 0.5699 
                & 0.7135 & 0.6996 
                & 0.6217 & 0.5703 
                & 0.6494 & 0.6307 
                & 0.6389 & 0.6176 \\
            \cellcolor{pastelPink} & SEEval
                & 0.6207 & 0.6146
                & 0.7128 & 0.7055
                & 0.6132 & 0.6139
                & 0.5941 & 0.5950
                & 0.6352 & 0.6323 \\
            \cellcolor{pastelPink} & CheckEval
                & 0.6269 & 0.6174 
                & 0.7215 & 0.7206 
                & 0.5806 & 0.5766 
                & 0.6512 & 0.6664 
                & \textbf{\phantom{0}0.6451$^\ast$} & \textbf{\phantom{0}0.6453$^\ast$} \\
            \cdashline{2-12}
            \cellcolor{pastelPink}\textbf{Qwen2.5-72B} & G-Eval 
                & 0.5650 & 0.5507 
                & 0.6944 & 0.6870 
                & 0.6122 & 0.6217 
                & 0.5927 & 0.5812 
                & \textbf{0.6161} & \textbf{0.6102} \\
            \cellcolor{pastelPink} & SEEval
                & 0.5448 & 0.5419
                & 0.6605 & 0.6552
                & 0.5942 & 0.6066
                & 0.5879 & 0.5896
                & 0.5969 & 0.5983 \\
            \cellcolor{pastelPink} & CheckEval
                & 0.5551 & 0.5506 
                & 0.7204 & 0.7199 
                & 0.4769 & 0.4873 
                & 0.6252 & 0.6398 
                & 0.5944 & 0.5994 \\
            \midrule
            \cellcolor{pastelBlue}\textbf{Mistral-Small} & G-Eval
                & 0.4439 & 0.4215 
                & 0.6550 & 0.6411 
                & 0.6939 & 0.5102 
                & 0.5103 & 0.4996 
                & \textbf{0.5758} & \textbf{0.5181} \\
            \cellcolor{pastelBlue} & SEEval
                & 0.2531 & 0.2478
                & 0.1768 & 0.1971
                & 0.1276 & 0.1238
                & 0.1169 & 0.1274
                & 0.1686 & 0.1740 \\
            \cellcolor{pastelBlue} & CheckEval
                & 0.3925 & 0.4225 
                & 0.6061 & 0.5914 
                & 0.4789 & 0.4826 
                & 0.4191 & 0.4777 
                & 0.4742 & 0.4935 \\
            \cdashline{2-12}
            \cellcolor{pastelBlue}\textbf{Gemma2-27B} & G-Eval 
                & 0.4086 & 0.4337 
                & 0.3286 & 0.2928 
                & 0.2680 & 0.2361 
                & 0.2173 & 0.1953 
                & 0.3056 & 0.2895 \\
            \cellcolor{pastelBlue} & SEEval
                & 0.4551 & 0.4621
                & 0.4212 & 0.4512
                & 0.3551 & 0.3795
                & 0.4627 & 0.4215
                & 0.4235 & 0.4286 \\
            \cellcolor{pastelBlue} & CheckEval
                & 0.5036 & 0.4952 
                & 0.6390 & 0.6323 
                & 0.3794 & 0.3718 
                & 0.5825 & 0.5714 
                & \textbf{0.5261} & \textbf{0.5177} \\
            \cdashline{2-12}
            \cellcolor{pastelBlue}\textbf{Qwen2.5-32B} & G-Eval
                & 0.4834 & 0.4515 
                & 0.3663 & 0.2697 
                & 0.4616 & 0.3082 
                & 0.5367 & 0.4924 
                & 0.4620 & 0.3804 \\
            \cellcolor{pastelBlue} & SEEval
                & 0.4551 & 0.4351
                & 0.4116 & 0.4642
                & 0.4531 & 0.3621
                & 0.5517 & 0.5921
                & 0.4679 & 0.4634 \\
            \cellcolor{pastelBlue} & CheckEval
                & 0.4918 & 0.4702 
                & 0.6914 & 0.6806 
                & 0.4139 & 0.4363 
                & 0.6300 & 0.6350 
                & \textbf{0.5568} & \textbf{0.5555} \\
            \midrule
            \cellcolor{pastelOlive}\textbf{Llama3.1-8B} & G-Eval
                & 0.1109 & 0.1013 
                & 0.1031 & 0.0813 
                & 0.1702 & 0.0959 
                & 0.0667 & 0.0765 
                & 0.1127 & 0.0887 \\
            \cellcolor{pastelOlive} & SEEval
                & 0.2531 & 0.2478
                & 0.1768 & 0.1971
                & 0.1276 & 0.1238
                & 0.1169 & 0.1274
                & 0.1686 & 0.1740 \\
            \cellcolor{pastelOlive}& CheckEval
                & 0.5046 & 0.4986 
                & 0.5200 & 0.5069 
                & 0.3972 & 0.3934 
                & 0.4050 & 0.3876 
                & \textbf{0.4567} & \textbf{0.4466} \\
            \cdashline{2-12}
            \cellcolor{pastelOlive}\textbf{Gemma2-9B} & G-Eval
                & 0.4357 & 0.3879 
                & 0.5512 & 0.4123 
                & 0.4742 & 0.3055 
                & 0.3681 & 0.2969 
                & 0.4573 & 0.3507 \\
            \cellcolor{pastelOlive} & SEEval
                & 0.4130 & 0.4303
                & 0.6116 & 0.6016
                & 0.4334 & 0.4441
                & 0.5020 & 0.5087
                & 0.4900 & 0.4962 \\
            \cellcolor{pastelOlive} & CheckEval
                & 0.3943 & 0.4232 
                & 0.6520 & 0.6588 
                & 0.4167 & 0.4136 
                & 0.4971 & 0.5137 
                & \textbf{0.4900} & \textbf{0.5023} \\
            \cdashline{2-12}
            \cellcolor{pastelOlive}\textbf{Qwen2.5-7B} & G-Eval
                & 0.4625 & 0.4540 
                & 0.5496 & 0.5111 
                & 0.3346 & 0.1429 
                & 0.4459 & 0.4421 
                & 0.4481 & 0.3875 \\
            \cellcolor{pastelOlive} & SEEval
                & 0.4130 & 0.3918
                & 0.5747 & 0.5735
                & 0.4681 & 0.4551
                & 0.4648 & 0.4322
                & 0.4802 & 0.4632 \\
            \cellcolor{pastelOlive} & CheckEval
                & 0.3704 & 0.3840 
                & 0.6329 & 0.6266 
                & 0.4712 & 0.4247
                & 0.4489 & 0.4486 
                & \textbf{0.4809} & \textbf{0.4710} \\
            \midrule
            \cellcolor{pastelLavender}\textbf{GPT-4 Turbo} & G-Eval
                & 0.4924 & 0.4719 
                & 0.7026 & 0.6900 
                & 0.6112 & 0.6126 
                & 0.5724 & 0.5512 
                & 0.5947 & 0.5814 \\
            \cellcolor{pastelLavender} & SEEval
                & 0.5012 & 0.5162
                & 0.7123 & 0.7221
                & 0.6232 & 0.6231
                & 0.5829 & 0.5922
                & 0.6049 & 0.6134 \\
            \cellcolor{pastelLavender} & CheckEval
                & 0.5209 & 0.5232 
                & 0.7367 & 0.7438 
                & 0.6292 & 0.6341 
                & 0.6425 & 0.6476 
                & \textbf{0.6323} & \textbf{0.6372} \\
            \cdashline{2-12}
            \cellcolor{pastelLavender}\textbf{GPT-4o} & G-Eval
                & 0.5917 & 0.5669 
                & 0.6111 & 0.5770 
                & 0.3903 & 0.1655 
                & 0.4770 & 0.4255 
                & 0.5175 & 0.4337 \\
            \cellcolor{pastelLavender} & SEEval
                & 0.6011 & 0.5881
                & 0.6551 & 0.5822
                & 0.4512 & 0.2620
                & 0.5331 & 0.4627
                & 0.5601 & 0.4738 \\
            \cellcolor{pastelLavender} & CheckEval
               & 0.5889 & 0.5790 
               & 0.7362 & 0.7354 
               & 0.5869 & 0.5761 
               & 0.6462 & 0.6448 
               & \textbf{0.6395} & \textbf{0.6338} \\
            \cdashline{2-12}
            \cellcolor{pastelLavender}\textbf{GPT-4o-mini} & G-Eval
                & 0.5424 & 0.5333 
                & 0.6024 & 0.5623 
                & 0.5748 & 0.5744 
                & 0.5977 & 0.5756 
                & 0.5793 & 0.5614 \\
            \cellcolor{pastelLavender} & SEEval
                & 0.5426 & 0.5277
                & 0.6051 & 0.5771
                & 0.5831 & 0.5651
                & 0.5441 & 0.5569
                & 0.5687 & 0.5567 \\
            \cellcolor{pastelLavender} & CheckEval
                & 0.5140 & 0.5171 
                & 0.5980 & 0.5984 
                & 0.6362 & 0.6241 
                & 0.6038 & 0.6160 
                & \textbf{0.5880} & \textbf{0.5889} \\
            \bottomrule
        \end{tabular}
    }
    \caption{Turn-level Spearman ($\rho$) and Pearson ($r$) correlations on Topical-Chat. The best score per model category is \textbf{bolded}, and the highest overall score is marked with \textbf{*}.}
    \label{tab:main-corr-result-topical-chat}
\end{table*}

\begin{table*}
    \centering
    \setlength{\tabcolsep}{10pt}
    \renewcommand{\arraystretch}{1.2}
    \resizebox{\textwidth}{!}{%
\label{tab:iea-summ}
\begin{tabular}{l l ll ll ll ll ll} 
\toprule
\multirow{2}{*}{
\begin{tabular}[c]{@{}l@{}}\textbf{Model}\\\textbf{Group}\end{tabular}
} & 
\multirow{2}{*}{
\begin{tabular}[c]{@{}l@{}}\textbf{Evaluation}\\\textbf{Methods}\end{tabular}
} & 
\multicolumn{2}{c}{\textbf{Coherence}} & 
\multicolumn{2}{c}{\textbf{Engagingness}} & 
\multicolumn{2}{c}{\textbf{Groundedness}} & 
\multicolumn{2}{c}{\textbf{Naturalness}} & 
\multicolumn{2}{c}{\textbf{Average}} \\
& & 
\multicolumn{1}{c}{$\alpha$} & 
\multicolumn{1}{c}{$\kappa$} & 
\multicolumn{1}{c}{$\alpha$} & 
\multicolumn{1}{c}{$\kappa$} & 
\multicolumn{1}{c}{$\alpha$} & 
\multicolumn{1}{c}{$\kappa$} & 
\multicolumn{1}{c}{$\alpha$} & 
\multicolumn{1}{c}{$\kappa$} & 
\multicolumn{1}{c}{$\alpha$} & 
\multicolumn{1}{c}{$\kappa$} \\
\midrule
All & G-Eval & 0.0651 & 0.3051 & 0.0418 & 0.3263 & 0.0825 & 0.4443 & 0.0462 & 0.2871 & 0.0589 & 0.3407\\
    & SEEval & 0.0741 & 0.3123 & 0.0668 & 0.3185 & 0.0674 & 0.3089 & 0.0717 & 0.3032 & 0.0700 & 0.3100 \\
    & CheckEval & 0.4796 & 0.4796 & 0.4354 & 0.4354 & 0.3995 & 0.3995 & 0.4830 & 0.4830 & 0.4494 & 0.4494 \\
\midrule
Large & G-Eval & 0.0070 & 0.4550 & 0.0110 & 0.5134 & 0.0030 & 0.7288 & 0.0371 & 0.3378 & 0.0145 & 0.5088 \\
    & SEEval & 0.5573 & 0.6091 & 0.5416 & 0.6074 & 0.5528 & 0.6137 & 0.5482 & 0.6085 & 0.5500 & 0.6100 \\
    & CheckEval & 0.6486 & 0.6486 & 0.6626 & 0.6626 & 0.6263 & 0.6263 & 0.7569 & 0.7569 & 0.6736 & 0.6736\\
\midrule
Medium & G-Eval & 0.1680 & 0.1361 & 0.0115 & 0.2581 & 0.0572 & 0.2907 & 0.0384 & 0.2074 & 0.0688 & 0.2231 \\
    & SEEval & 0.0527 & 0.3426 & 0.0614 & 0.3362 & 0.0595 & 0.3407 & 0.0659 & 0.3393 & 0.0600 & 0.3400 \\
    & CheckEval & 0.3635 & 0.3635 & 0.5338 & 0.5338 & 0.4486 & 0.4486 & 0.6715 & 0.6715 & 0.5044 & 0.5043 \\
\midrule
Small & G-Eval & 0.0357 & 0.1535 & 0.0287 & 0.1528 & 0.0603 & 0.2139 & 0.0242 & 0.1343 & 0.0372 & 0.1636\\
    & SEEval & 0.1615 & 0.1487 & 0.1553 & 0.1511 & 0.1628 & 0.1494 & 0.1674 & 0.1542 & 0.1600 & 0.1500 \\
    & CheckEval & 0.4040 & 0.4040 & 0.2127 & 0.2127 & 0.0218 & 0.0218 & 0.0289 & 0.0289 & 0.1669 & 0.1668\\
\midrule
GPT & G-Eval & 0.0079 & 0.4970 & 0.0698 & 0.3936 & 0.0225 & 0.6910 & 0.0536 & 0.4067 & 0.0385 & 0.4971\\
    & SEEval & 0.1191 & 0.5057 & 0.1217 & 0.5123 & 0.1175 & 0.5148 & 0.1283 & 0.5021 & 0.1200 & 0.5100 \\
    & CheckEval & 0.5651 & 0.5651 & 0.2452 & 0.2452 & 0.6124 & 0.6124 & 0.7352 & 0.7352 & 0.5395 & 0.5395\\
\midrule
Top-3 & G-Eval & 0.0234 & 0.4389 & 0.0015 & 0.6510 & 0.0020 & 0.7701 & 0.0752 & 0.3773 & 0.0255 & 0.5593\\
    & SEEval & 0.0597 & 0.3375 & 0.0614 & 0.3401 & 0.0558 & 0.3418 & 0.0579 & 0.3405 & 0.0600 & 0.3400 \\
    & CheckEval & 0.6215 & 0.6215 & 0.2481 & 0.2480 & 0.6435 & 0.6434 & 0.7813 & 0.7812 & 0.5736 & 0.5736\\
\bottomrule
\end{tabular}
    }
    \caption{IEA - Topical-Chat.}
    \label{tab:iea-full result-topicalchat}
\end{table*}

\begin{figure*}[ht]
    \centering
    \begin{tcolorbox}[
    colback=blue!10!white,       
    colframe=black,             
    colbacktitle=blue!50!white, 
    coltitle=black,             
    rounded corners,            
    boxrule=0.8mm,              
    width=\textwidth,           
    fonttitle=\large,
    title={\textbf{Evaluation Prompt for SummEval}}, 
    ]
        \textbf{<Task Overview>}\\
        \texttt{Your task is to read a provided news article and its summary, then answer `yes’ or `no’ to specific questions. These questions will relate to a particular dimension of the summary.}

        \textbf{\\<dimension Definition>}\\
        \texttt{\textless dimension\textgreater - \textless definition\textgreater}

        \textbf{\\<Instructions>}
        \begin{enumerate}
            \item \texttt{Read these instructions thoroughly.}
            \item \texttt{Carefully read both the Article and the Summary.}
            \item \texttt{Understand the given questions and the definition of the \textless dimension\textgreater.}
            \item \texttt{Respond to each question with `yes’ or `no’. Base your answers on a clear rationale.}
            \item \texttt{Follow the specified format for your answers.}
        \end{enumerate}

        \textbf{\\<Answer Format>}\\
        \texttt{Q1: [Your Answer]\\
        Q2: [Your Answer]\\
        ...}

        \textbf{\\\# Article \#}\\
        \texttt{\textless source\textgreater}

        \textbf{\\\# Summary \#}\\
        \texttt{\textless summary\textgreater}

        \textbf{\\\# Questions \#}\\
        \texttt{\textless questions\textgreater}

        \textbf{\\\# Response \#}\\
        \texttt{Provide your answers to the given questions, following the specified Answer Format.}
    \end{tcolorbox}
    \caption{Evaluation Prompt - SummEval}
    \label{fig:evaluation_prompt_summeval}
\end{figure*}

\begin{figure*}[ht]
    \centering
    \begin{tcolorbox}[
    colback=yellow!10!white,      
    colframe=black,              
    colbacktitle=yellow!50!white,
    coltitle=black,              
    rounded corners,             
    boxrule=0.8mm,               
    width=\textwidth,            
    fonttitle=\large,
    title={\textbf{Evaluation Prompt for Topical-Chat}}, 
    ]
        \textbf{<Task Overview>}\\
        \texttt{You will be given a conversation between two individuals. You will then be given one potential response for the next turn in the conversation. The response concerns an interesting fact, which will be provided as well.\\
        Your task is to read a provided conversation history, corresponding fact, and response, then answer `yes’ or `no’ to specific questions. These questions will relate to a particular dimension of the response.}

        \textbf{\\<dimension Definition>}\\
        \texttt{\textless dimension\textgreater - \textless definition\textgreater}

        \textbf{\\<Instructions>}
        \begin{enumerate}
            \item \texttt{Read these instructions thoroughly.}
            \item \texttt{Carefully read the Conversation History, the Corresponding Fact, and the Response.}
            \item \texttt{Understand the given questions and the definition of the \textless dimension\textgreater.}
            \item \texttt{Respond to each question with `yes’ or `no’. Base your answers on a clear rationale.}
            \item \texttt{Follow the specified format for your answers.}
        \end{enumerate}

        \textbf{\\<Answer Format>}\\
        \texttt{Q1: [Your Answer]\\
        Q2: [Your Answer]\\
        ...}

        \textbf{\# Conversation History \#}\\
        \texttt{\textless document\textgreater}

        \textbf{\\\# Corresponding Fact \#}\\
        \texttt{\textless fact\textgreater}

        \textbf{\\\# Response \#}\\
        \texttt{\textless response\textgreater}

        \textbf{\\\# Questions \#}\\
        \texttt{\textless questions\textgreater}

        \textbf{\\\# Your Answer \#}\\
        \texttt{Provide your answers to the given questions, following the specified Answer Format.}
    \end{tcolorbox}
    \caption{Evaluation Prompt - Topical-Chat}
    \label{fig:evaluation_prompt_topicalchat}
\end{figure*}

\begin{figure*}[ht!]
    \centering
    \begin{tcolorbox}[
    colback=red!10!white,      
    colframe=black,             
    colbacktitle=red!50!white,
    coltitle=black,             
    rounded corners,            
    boxrule=0.8mm,              
    width=\textwidth,           
    fonttitle=\large,
    title={\textbf{Augmentation - Question Diversification Prompt}}, 
    ]
        \textbf{<Task Overview>}\\
        \texttt{You will be provided with: 1) Information about the benchmark to be evaluated, 2) The main concept being assessed in the benchmark, and 3) Seed questions that include key components and sub-questions related to this concept.\\
        Your task is to create additional sub-questions for the key components to comprehensively assess the main concept. Each sub-question must meet given conditions to ensure a high-quality question set.}
        
        \textbf{\\1) Benchmark Information:}\\
        \texttt{\{benchmark description\}}
        
        \textbf{\\2) Main Concept in the Benchmark:}\\
        \texttt{\{concept\}: \{description\}}
        
        \textbf{\\3) Key Components and Seed Questions:}\\
        \texttt{\{seed questions\}}
        
        \textbf{\\<Conditions for a Good Question List>}
        \texttt{\\\{conditions\}}

        \textbf{\\<Constraints>}\\
        \texttt{- Each sub-question must be answerable with a simple `yes’ or `no’.\\
        - A `yes’ answer should indicate that the sentence improves the specified evaluation criterion (e.g., Coherence, Relevance).\\
        - Each question should assess only a single dimension or concept.\\
        - Each question should not ask about more than one topic or concept.}
    \end{tcolorbox}
    \caption{Augmentation - Question Diversification Prompt}
    \label{fig:augmentation_prompt_diversification}
\end{figure*}

\begin{figure*}
    \centering
    \begin{tcolorbox}[
        colback=green!10!white,      
        colframe=black,             
        colbacktitle=green!50!white,
        coltitle=black,             
        rounded corners,            
        boxrule=0.8mm,              
        width=\textwidth,           
        fonttitle=\large,
        title=\textbf{Augmentation - Question Elaboration Prompt} 
    ]
    \textbf{<TASK OVERVIEW>}\\
    \texttt{Your task is to generate multiple additional questions to evaluate benchmark performance under specific constraints. You will receive the key component and sub-component evaluating \{dimension\} and the question related to it. The definition of \{dimension\} is as follows: \{def\}. The evaluation for dimension \{dimension\} will be centered around the key component \{key components\}.}

    \textbf{\\<TASK>}\\
    \texttt{\# Your role: You have to break down sub-questions into 3 to 10 sub-sub-questions considering \{dimension\} when pairs of seed name and question are given.\\
    \# Benchmark information: \{benchmark info\}}

    \textbf{\\<CONSTRAINTS>}\\
    \texttt{\{constraints\}}

    \textbf{\\<Conditions for a Good Question List>}\\
    \texttt{\{conditions\}}

    \textbf{\\<FORMAT>}\\
    \texttt{
    1. sub\_component\_name\_1:\\
       1-1. q1-1\_origin\_question\\
       1-1-1. q1-1-1\_aug\_question\\
       1-1-2. q1-1-2\_aug\_question\\
       ...\\
       1-2. q1-2\_origin\_question\\
       1-2-1. q1-2-1\_aug\_question\\
       1-2-2. q1-2-2\_aug\_question\\
       ...\\
    \\2. sub\_component\_name\_2:\\
       2-1. q2-1\_origin\_question\\
       2-1-1. q2-1-1\_aug\_question\\
       ...\\
       2-2. q2-2\_origin\_question\\
       ...}

    \textbf{\\<EXAMPLE>}\\
    \texttt{\{example\}}
    \end{tcolorbox}
    \caption{Augmentation - Question Elaboration Prompt}
    \label{fig:augmentation_prompt_elaboration}
\end{figure*}

\begin{figure*}[ht]
    \centering
    \begin{tcolorbox}[
    colback=orange!10!white,      
    colframe=black,             
    colbacktitle=orange!50!white,
    coltitle=black,             
    rounded corners,            
    boxrule=0.8mm,              
    width=\textwidth,           
    fonttitle=\large,
    title={\textbf{Filtering Prompt}}, 
    ]
        \textbf{<Task Overview>}\\
        \texttt{Your task is to filter out questions from a list based on the following criteria:}
        
        \textbf{\\1) dimension Alignment:}\\
        \texttt{- dimension definition: \{dimension def\}\\
        - Remove questions that deviate from the given dimension's definition.\\
        - Remove questions that are more closely related to other dimensions than the current one.}

        \textbf{\\2) Redundancy:}\\
        \texttt{- Remove questions that:\\
        \hspace{1em}* Ask for the same or very similar information (even if phrased differently).\\
        \hspace{1em}* Convey very similar meanings without adding unique insight.}

        \textbf{\\3) Style:}\\
        \texttt{- Remove questions that:\\
        \hspace{1em}* Use overly exaggerated wording.\\
        \hspace{1em}* Focus on excessively detailed or minor points that don't meaningfully affect overall quality.}

        \textbf{\\4) Benchmark Context}\\
        \texttt{- Name: Topical-Chat\\
        - Purpose: Evaluation of knowledge-grounded dialogue systems\\
        - Key Metrics: Naturalness, Coherence, Engagingness, Groundedness\\
        - Do not modify any of the remaining questions or generate new ones.\\
        - Keep questions in their original dictionary format.}

        \textbf{\\5) Sub-dimensions and Questions:}\\
        \texttt{\{format\_sub\_dimensions(sub\_dimensions)\}}

        \textbf{\\6) Output Requirements:}\\
        \texttt{- Output format: JSON only\\
        - Structure:}
        \begin{verbatim}
{"Sub-dimension Name": [
    "Filtered Question 1",
    "Filtered Question 2"]}
        \end{verbatim}
        \textbf{<Important Note>}\\
        \texttt{
        - Do not modify the content of remaining questions\\
        - Do not generate new questions\\
        - Maintain the original dictionary format\\
        - Only remove questions that fail the above criteria\\
        - Do not remove entire sub-dimensions or their keys unless no valid questions remain.}
    \end{tcolorbox}
    \caption{Filtering Prompt}
    \label{fig:filtering_prompt}
\end{figure*}

\newpage

\begin{table*}[ht!]
    \centering
    \setlength{\tabcolsep}{8pt} 
    \renewcommand{\arraystretch}{1.4} 
    \resizebox{0.95\textwidth}{!}{%
        \footnotesize
        \begin{tabular}{p{3cm} !{\vrule} p{4cm} !{\vrule} p{9cm}}
            \toprule
            \textbf{Dimension} & \textbf{Sub-dimension} & \textbf{Seed Questions} \\
            \midrule
            \multirow{3}{*}{\textbf{Coherence}} 
            & Topic Maintenance & Does the summary consistently focus on the central topic without deviating into unrelated areas? \\
            \cline{2-3}
            & Logical Flow & Does the summary present information in a logical order? \\
            \cline{2-3}
            & Consistent Point of View & Is the point of view or perspective in the summary consistent with the source? \\
            \midrule
            \multirow{3}{*}{\textbf{Consistency}} 
            & Factual Consistency & Does the summary accurately represent the facts from the source? \\
            \cline{2-3}
            & No New Information & Does the summary avoid introducing information not present in the original source? \\
            \cline{2-3}
            & Contextual Accuracy & Does the summary preserve the original purpose or intent of the source document? \\
            \midrule
            \multirow{4}{*}{\textbf{Fluency}} 
            & Formatting & Is the summary free from formatting issues and correctly capitalized throughout? \\
            \cline{2-3}
            & Grammar & Are all sentences grammatically correct and free from errors? \\
            \cline{2-3}
            & Completeness & Are all sentences complete, with no fragments or missing components? \\
            \cline{2-3}
            & Readability & Is the summary easy to read, without unnecessary complexity? \\
            \midrule
            \multirow{5}{*}{\textbf{Relevance}} 
            & Content Coverage & Does the summary encapsulate all critical points of the source document? \\
            \cline{2-3}
            & Topic Consistency & Does the summary maintain the main topic of the source? \\
            \cline{2-3}
            & Consistent Use of Terminology & Does the summary use the same terminology or jargon as the source? \\
            \cline{2-3}
            & Use of Key Terms and Phrases & Does the summary incorporate key terms and phrases from the source material effectively? \\
            \cline{2-3}
            & Importance & Is each point mentioned in the summary important to the overall understanding of the original text? \\
            \bottomrule
        \end{tabular}
    }
    \caption{Dimensions, sub-dimensions, and corresponding seed questions for SummEval.}
    \label{tab:summary-dimensions}
\end{table*}

\begin{table*}[t]
    \centering
    \setlength{\tabcolsep}{8pt} 
    \renewcommand{\arraystretch}{1.4} 
    \resizebox{0.95\textwidth}{!}{%
        \footnotesize
        \begin{tabular}{p{3cm} !{\vrule} p{4cm} !{\vrule} p{9cm}}
            \toprule
            \textbf{Dimension} & \textbf{Sub-dimension} & \textbf{Seed Questions} \\
            \midrule
            \multirow{2}{*}{\textbf{Coherence}} 
            & Logical Flow & Does the response logically follow from the earlier part of the conversation, maintaining a clear flow of ideas? \\
            \cline{2-3}
            & Relevance & Is the response directly relevant to the content and context of the previous dialogue? \\
            \cline{2-3}
            & \multirow{2}{*}{Continuity} 
            & Does the response stay consistent with the topic discussed in the previous dialogue? \\
            & & Does the response integrate smoothly with the ongoing conversation, ensuring a coherent progression? \\
            \midrule
            \multirow{3}{*}{\textbf{Engagingness}} 
            & Informative & Does the response add meaningful value to the conversation? \\
            \cline{2-3}
            & Emotional Engagement & Is the response friendly, polite, and empathetic? \\
            \cline{2-3}
            & \multirow{2}{*}{Interest Level} 
            & Does the response capture interest or intrigue, making the conversation more engaging? \\
            & & Does the response actively contribute to keeping the conversation lively and engaging? \\
            \midrule
            \multirow{3}{*}{\textbf{Groundedness}} 
            & \multirow{3}{*}{Relevance} 
            & Does the response appropriately address the preceding question or statement? \\
            & & Does the answer provide new information while maintaining the flow of the conversation? \\
            & & Does it effectively utilize the key information that has been mentioned in the conversation? \\
            \cline{2-3}
            & \multirow{2}{*}{Consistency} 
            & Does the response remain consistent with previous utterances? \\
            & & Does it avoid contradicting previously provided information? \\
            \midrule
            \multirow{5}{*}{\textbf{Naturalness}} 
            & Avoid repetition & Does the response avoid unnecessary repetition of the same content between sentences? \\
            \cline{2-3}
            & Context relevance & Are all the sentences relevant to the topic of conversation and used naturally within the context? \\
            \cline{2-3}
            & Clarity & Is the overall message clear and easy to understand? \\
            \cline{2-3}
            & \multirow{2}{*}{Word choice and tone} 
            & Is the tone consistent throughout? \\
            & & Are there no major grammatical errors? \\
            \bottomrule
        \end{tabular}
    }
    \caption{Dimensions, sub-dimensions, and corresponding seed questions for Topical-Chat.}
    \label{tab:dialogue-dimensions}
\end{table*}